\documentclass[%
 aip,
 jmp,%
 amsmath,amssymb,
reprint,%
]{revtex4-2}
\usepackage{fullpage}
\usepackage{siunitx}
\usepackage{subfigure}
\usepackage{natbib}
\usepackage{graphicx}
\usepackage{dcolumn}
\usepackage{bm}

\begin{document}

\preprint{Physics of Fluids}

\title{Modelling pressure-Hessian from local velocity gradients information in an incompressible turbulent flow field using deep neural networks}

\author{Nishant Parashar}
 \email{nishantparashar14@gmail.com}
\affiliation{Department of Applied Mechanics, Indian Institute of Technology, Delhi, New Delhi, 110016, India}%

\author{Sawan S. Sinha}%
 \email{sawan@am.iitd.ac.in}
\affiliation{Department of Applied Mechanics, Indian Institute of Technology, Delhi, New Delhi, 110016, India}%

\author{Balaji Srinivasan}%
 \email{balaji.srinivasan@gmail.com}
\affiliation{Department of Mechanical Engineering, Indian Institute of Technology, Madras, Chennai, 600036, India}%

\date{\today}
\begin{abstract}
The understanding of the dynamics of the velocity gradients in turbulent flows is critical to understanding various non-linear turbulent processes. The pressure-Hessian and the viscous-Laplacian govern the evolution of the velocity-gradients and are known to be non-local in nature. Over the years, several simplified dynamical models have been proposed that models the viscous-Laplacian and the pressure-Hessian primarily in terms of local velocity gradients information. These models can also serve as closure models for the Lagrangian PDF methods. The recent fluid deformation closure model (RFDM) has been shown to retrieve excellent one-time statistics of the viscous process. However, the pressure-Hessian modelled by the RFDM has various physical limitations. In this work, we first demonstrate the limitations of the RFDM in estimating the pressure-Hessian. Further, we employ a tensor basis neural network (TBNN) to model the pressure-Hessian from the velocity gradient tensor itself. The neural network is trained on high-resolution data obtained from direct numerical simulation (DNS) of isotropic turbulence at Reynolds number of 433 (JHU turbulence database, JHTD). The predictions made by the TBNN are tested against two different isotropic turbulence datasets at Reynolds number of 433 (JHTD) and 315 (UP Madrid turbulence database, UPMTD) and channel flow dataset at Reynolds number of 1000 (UT Texas and JHTD). The evaluation of the neural network output is made in terms of the alignment statistics of the predicted pressure-Hessian eigenvectors with the strain-rate eigenvectors for turbulent isotropic flow as well as channel flow. Our analysis of the predicted solution leads to the discovery of ten unique coefficients of the tensor basis of strain-rate and rotation-rate tensors, the linear combination over which accurately captures key alignment statistics of the pressure-Hessian tensor.

\end{abstract}

\maketitle


\section{Introduction}
In a turbulent flow, various processes like energy cascade, intermittency, fluid element deformation are strongly related to the small scale velocity gradient field. Various experimental, direct numerical simulation and simple dynamical models based studies have been performed to understand the dynamics of the velocity gradient tensor \cite[]{luthi2005lagrangian, ashurst1987alignment, vieillefosse1982local,cantwell1992exact}. In continuation to these works, several other studies have been reported as well \cite[]{ashurst1987alignment, ashurst1987pressure, girimaji1990diffusion, ohkitani1993eigenvalue, pumir1994numerical, girimaji1995modified, o2005relationship, chevillard2006lagrangian, da2008invariants, chevillard2011lagrangian, soria1994study, pirozzoli2004direct_b, suman2012velocity, wang2012flow, vaghefi2015local, danish2016influence, parashar2017, parasharJFM, parasharIJHFF}. The pressure-Hessian and the viscous tensor are the two important processes governing the evolution of the velocity gradient tensor. These processes are inherently non-local in nature and are unclosed from a mathematical viewpoint. \citet{chevillard2008} developed a recent fluid deformation closure model (RFDM) for modelling the viscous tensor and the pressure-Hessian. Although the RFD model robustly captures various one-time statistics of the viscous tensor, it has various inherent limitations in predicting the pressure-Hessian tensor (discussed in section \ref{s:rfdm}). Hence, in this paper, we focus on modelling the pressure-Hessian tensor using velocity gradients information. Recently an improved model$-$recent fluid deformation of Gaussian fields (RFDG model) has been proposed by \citet{johnson2016}. It is an improvement over the RFD model in terms of predicting various one-time statistics of the velocity gradient tensor. However, the authors \cite[]{johnson2016} did not focus on any relevant statistics of the pressure-Hessian tensor. Due to this reason, in this work, all our comparisons will be made against the RFD model of \citet{chevillard2008}.

In the recent past, machine learning has gained popularity in the turbulence research community. The earliest such contribution in the field of machine learning aided turbulence research was made by \citet{duraisamy2014}, where the authors developed an intermittency transport-based model for bypass transition using machine learning and inverse modelling. Since then, a large number of researchers have tried to model various turbulence processes using machine learning models \cite[]{duraisamy2015, brendan2015, brendan2015b, parish2016,zhang2015,jack2017,duraisamy2019}. \citet{ling2016} employed a deep neural network to directly model the Reynolds stress anisotropy tensor using strain-rate and rotation-rate tensors. In doing so, they developed a novel tensor basis neural network (TBNN), which can be employed to map a given tensor from known input tensors. The TBNN has been shown to achieve superior performance by embedding tensor invariance properties in the network itself. Later \citet{fang2018} used the TBNN for turbulent channel flow and compared their results against standard turbulence models. \citet{sotgiu2018} developed a new framework in conjunction with TBNN for predicting turbulent heat fluxes. Further, \citet{geneva2019} developed a Bayesian tensor basis neural network for predicting the Reynolds stress anisotropy tensor.

As mentioned earlier, the recent fluid deformation closure model (RFDM) \cite[]{chevillard2008} is considered to be the state of the art model for pressure-Hessian calculation. However, the pressure-Hessian predicted by the RFD model shows nonphysical alignment tendencies with the strain rate tensor (explained in section \ref{s:rfdm}). Any further improvement in the existing model may require a deeper understanding of the complex relationship between pressure-Hessian and velocity gradients. For this task, we employ deep learning, which can potentially decipher any functional relationship that can potentially exist between the quantities of interest. The tensor basis neural network (TBNN) developed by \citet{ling2016} has already been shown to map tensorial quantities robustly. In this work, we use high resolution incompressible isotropic turbulence data from John Hopkins University turbulence database, JHTD \cite[]{JHUTD_1, JHUTD_2} (\url{http://turbulence.pha.jhu.edu}) to train a neural network model inspired by TBNN. Further, we show that by appropriate normalization of the input data and a few modifications in the network can lead to significant improvements in alignment characteristics of the predicted output. The predictions made by the TBNN are compared against two different isotropic turbulence datasets that were not used for training the network$-$(i) Taylor Reynolds number of 433, JHTB \cite[]{JHUTD_1, JHUTD_2} and (ii) isotropic turbulence at Taylor Reynolds number of 315 (UP Madrid database, \url{https://torroja.dmt.upm.es/turbdata/Isotropic}) \cite[]{cardesa2017}. To demonstrate the generality of the predicted solution in terms of alignment statistics for other types of flows, we also test the trained model for channel flow data at friction velocity of 1000 (UT Austin and JHU turbulence database) \cite[]{JHUTD_3}. Further evaluation of the neural network output helps us retrieve ten unique coefficients of the tensor basis of strain-rate and rotation-rate tensors, the linear combination over which can be used to predict the pressure-Hessian tensor robustly.

This paper is organized into six sections. In section \ref{s:goveq} we present the governing equations. In section \ref{s:rfdm}, we explain the limitations of the RFD model. In section \ref{s:NN}, we present the details of the tensor basis neural network architecture employed for this study. The analysis of the predicted solution from the TBNN is also presented in section \ref{s:NN}. Further, in section \ref{s:NN_mod}, we explain the modifications incorporated in the TBNN network and compare its results against state of the art RFD model. 
Section \ref{s:summary} concludes the paper with a brief summary.

\section{Governing Equations}
\label{s:goveq}
The governing equations of an incompressible flow field comprises of the continuity, momentum and state equation of a perfect gas:
\begin{align}
\frac{\partial{V_k}}{\partial{x_k}}&=0;
\label{eq:mass_con}\\
\frac{\partial{V_i}}{\partial{t}}+V_k\frac{\partial{V_i}}{\partial{x_k}}&=-\frac{1}{\rho}\frac{\partial{p}}{\partial{x_i}}+\frac{\mu}{\rho}\frac{\partial^2 V_i}{\partial{x_k}\partial{x_k}};
\label{eq:moment_con}\\
p&=\rho RT
\label{eq:state}
\end{align}
where $V_i$ and $x_i$ represents the velocity and position respectively. Density, pressure and temperature are represented by $\rho$ , $p$ and $T$, while $R$ denotes the gas constant.  The velocity gradient tensor is defined as:
 $$A_{ij} \equiv \frac{\partial{V_i}}{\partial{x_j}}.$$
Taking the gradient of momentum equation (\ref{eq:moment_con}), the exact evolution equation of $A_{ij}$ can be derived:
\begin{align}  
   \frac{DA_{ij}}{Dt}=-A_{ik}A_{kj} - \underbrace{\frac{\partial^2p}{\partial{x_i}\partial{x_j}}}_{\mathcal{P}_{ij}} + \underbrace{\nu\frac{\partial^2A_{ij}}{\partial{x_k}\partial{x_k}}}_{\Upsilon_{ij}}
   \label{eq:exact_evolA}
\end{align}
where $\boldsymbol{\mathcal{P}}$ and $\boldsymbol{\Upsilon}$  represent the pressure-Hessian and the viscous Laplacian governing the evolution of the velocity gradient tensor. The rate of change of $A_{ij}$ following a fluid particle is represented using the substantial derivative: $D / Dt \left(\equiv \partial / \partial{t} + V_{k} \partial / \partial{x_k} \right)$.

\section{Limitations of the RFD model for pressure-Hessian calculation}
\label{s:rfdm}
The state of the art model for pressure-Hessian calculation is the recent fluid deformation closure model (RFDM) developed by \citet[]{chevillard2008}. The RFD pressure-Hessian ($\boldsymbol{\mathcal{P}^{RFD}}$) is expressed as:
\begin{equation}
    \boldsymbol{\mathcal{P}^{RFD}} = -\frac{\{\boldsymbol{A}^2\}}{\{\boldsymbol{C_{\tau _k}^{-1}}\}} \boldsymbol{C_{\tau_k}^{-1}},
\end{equation}
where, $\boldsymbol{C}$ is the right Cauchy Green tensor modelled as: $\boldsymbol{C_{\tau k}}=e^{\tau_k \boldsymbol{A}}e^{\tau_k \boldsymbol{A^T}}$ and the symbol $\{\}$ represents the trace of the tensor. 
The pressure-Hessian predicted by the RFD model has some inherent inconsistencies as compared to the actual pressure-Hessian obtained from DNS. These limitations are listed below:
\begin{enumerate}
    \item \textit{$\boldsymbol{\mathcal{P}^{RFD}}$ is always positive-definite:} 
    It is evident that $\boldsymbol{C_{\tau k}}$ is a positive-definite matrix as it is basically a product of a real matrix ($e^{\tau_k \boldsymbol{A}}$) and it's transpose. Since the inverse of a positive-definite matrix is also positive-definite, therefore, $\boldsymbol{C_{\tau_k}^{-1}}$ is always positive-definite and $\boldsymbol{\mathcal{P}^{RFD}}$ is guaranteed to be either positive-definite or negative-definite, depending on the sign of $-\frac{\{\boldsymbol{A}^2\}}{\{\boldsymbol{C_{\tau _k}^{-1}}\}}$. Therefore, the eigenvalues of $\boldsymbol{\mathcal{P}^{RFD}}$ are either all negative or all positive. This behavior of $\boldsymbol{\mathcal{P}^{RFD}}$ is nonphysical, since the governing equations does not impose any such restriction on $\boldsymbol{\mathcal{P}}$ to be either positive-definite or negative-definite. $\boldsymbol{\mathcal{P}}$ is real symmetric by nature and hence will have at-least one positive and one negative eigenvalue most of the time. 

    \begin{figure}[h]
    \centering
    \subfigure[]{
    \includegraphics[width=4.5cm]{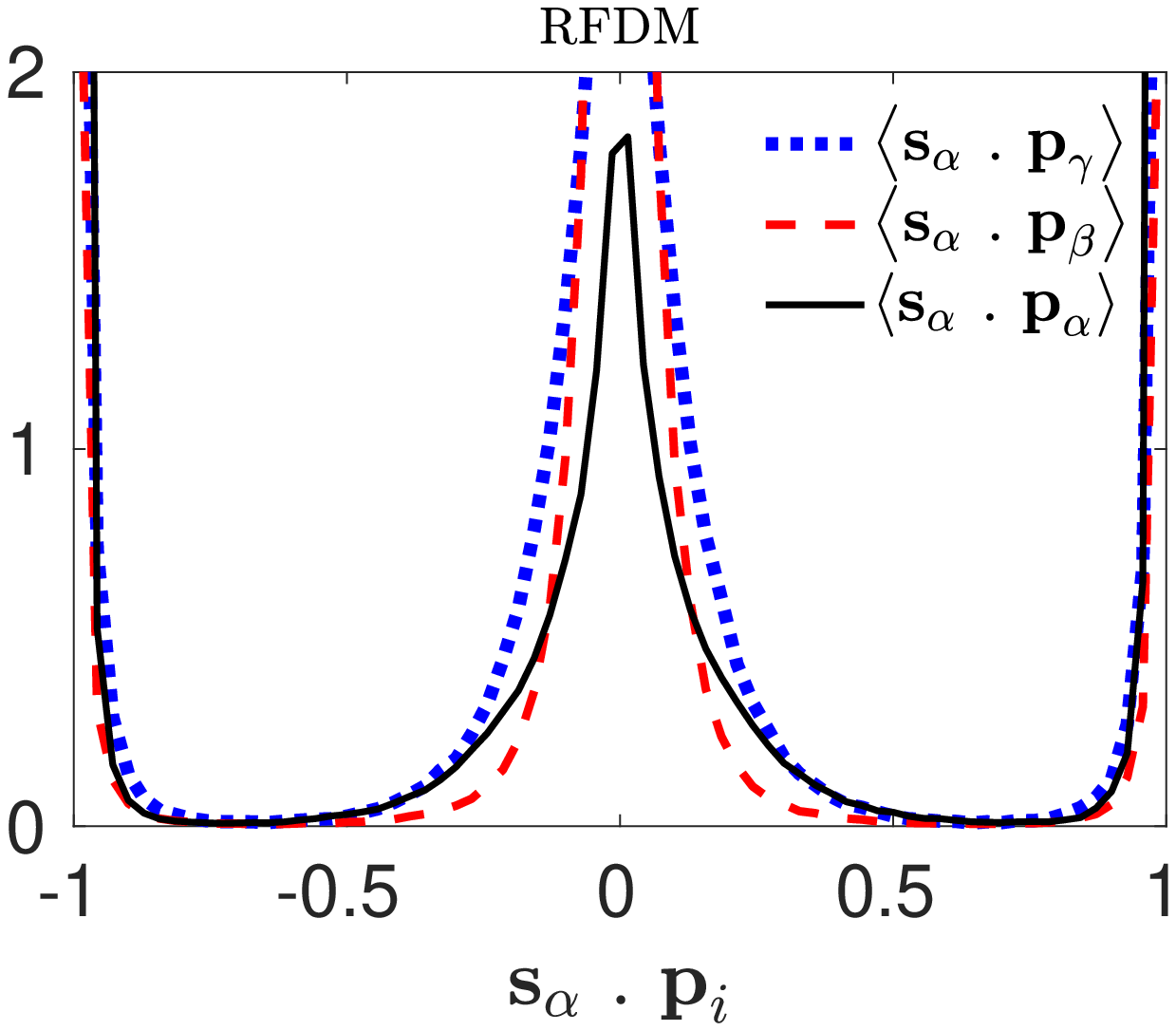}}
    \subfigure[]{
    \includegraphics[width=4.5cm]{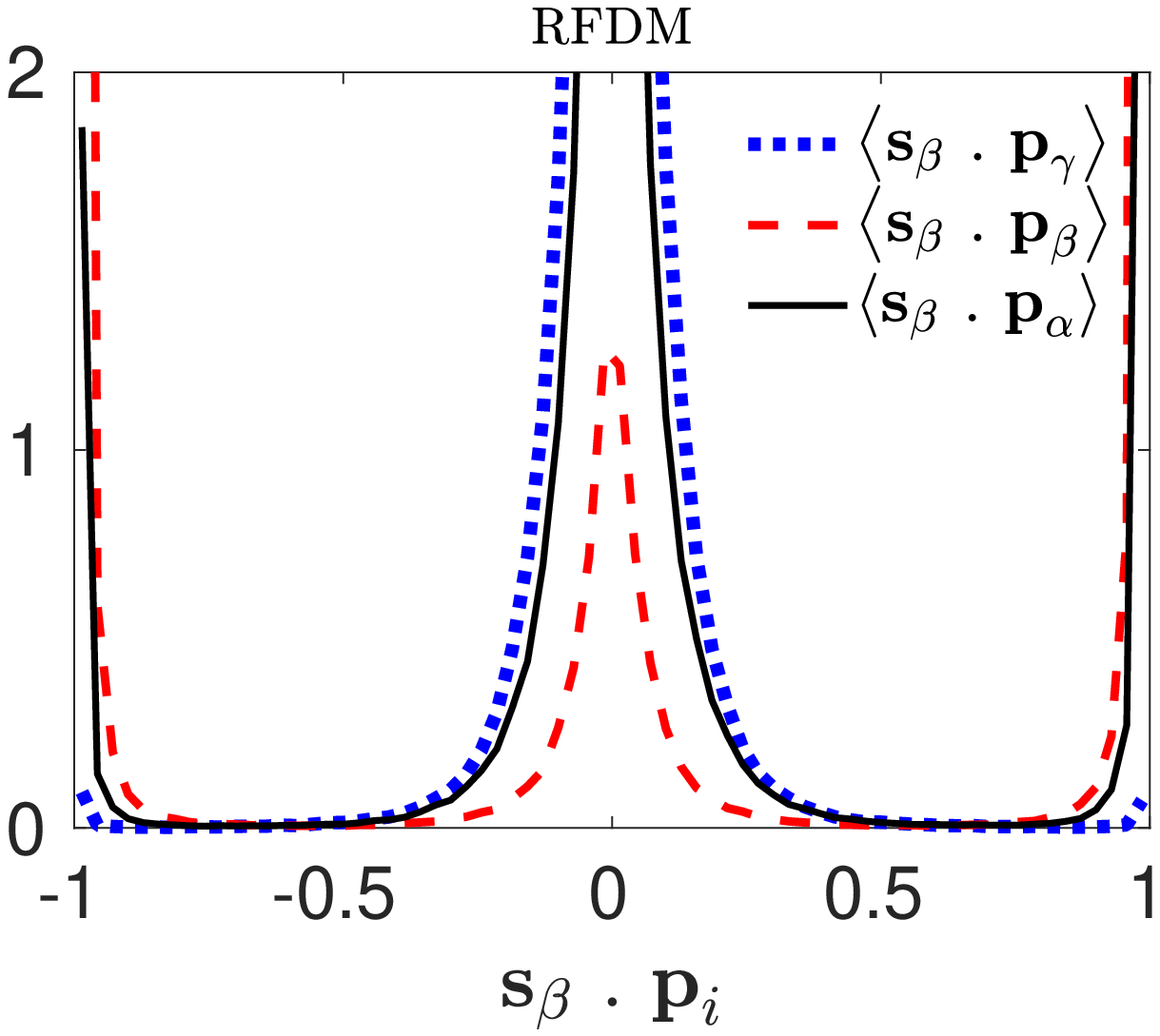}}
    \subfigure[]{
    \includegraphics[width=4.5cm]{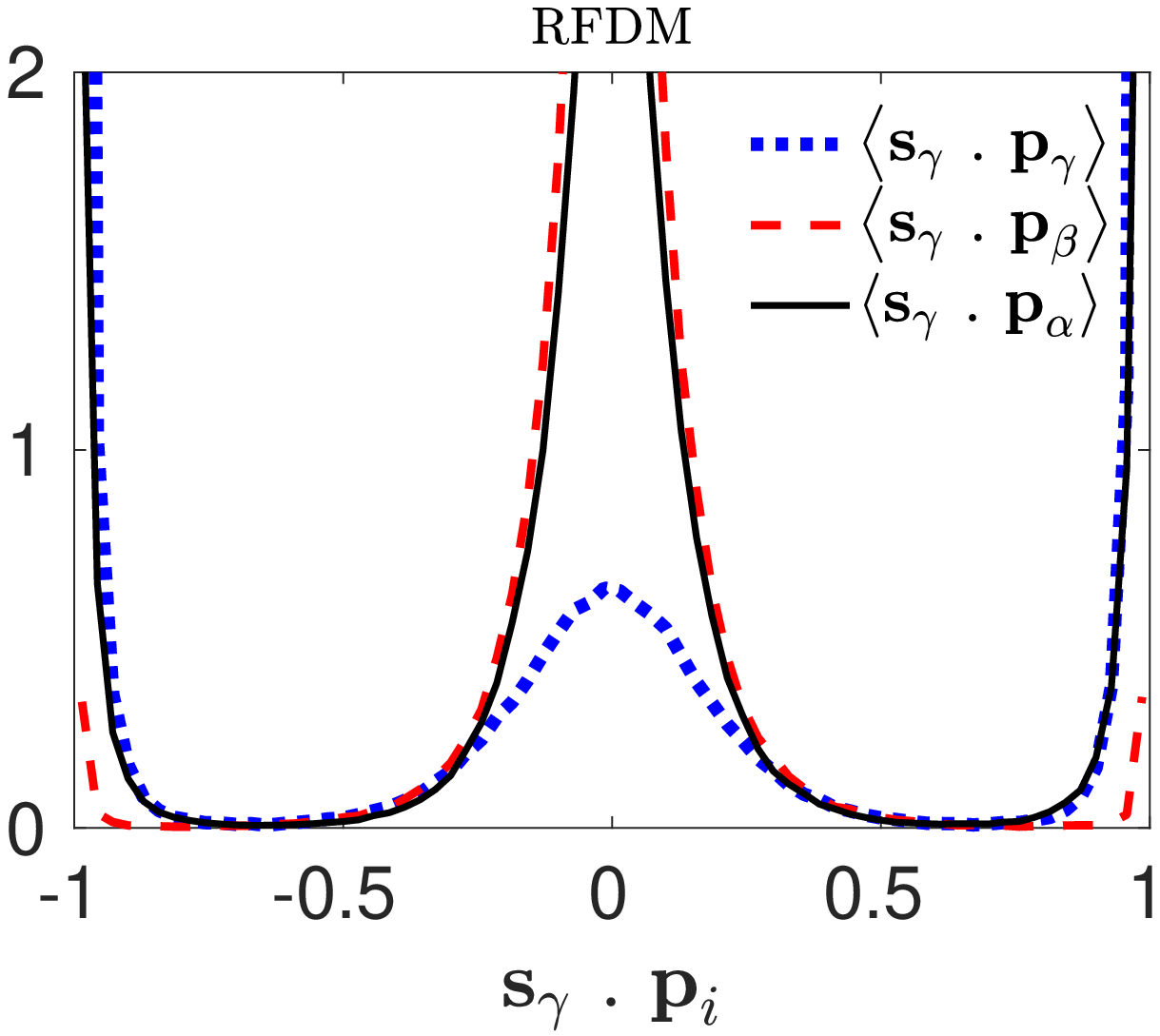}}
    \subfigure[]{
    \includegraphics[width=4.5cm]{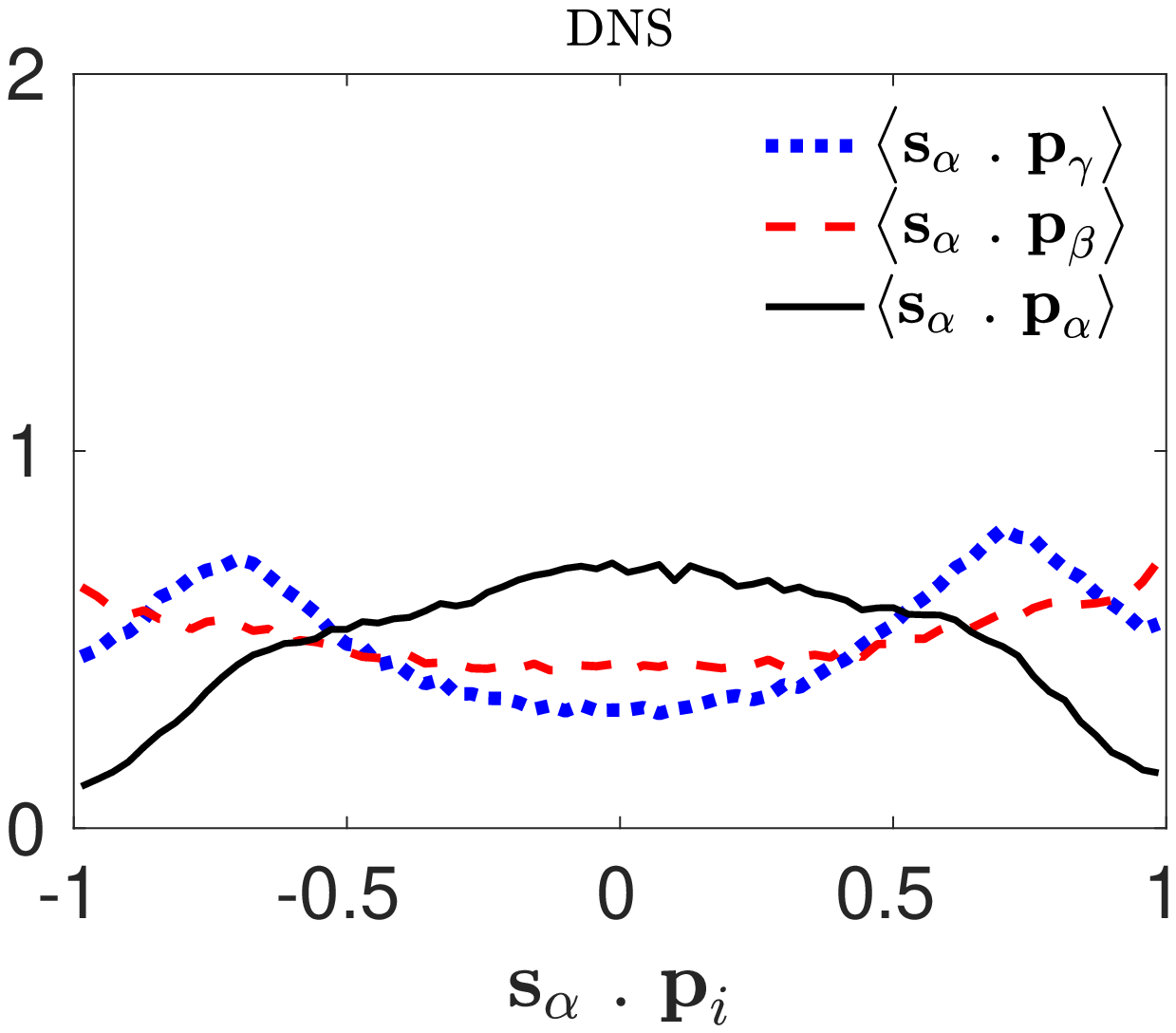}}
    \subfigure[]{
    \includegraphics[width=4.5cm]{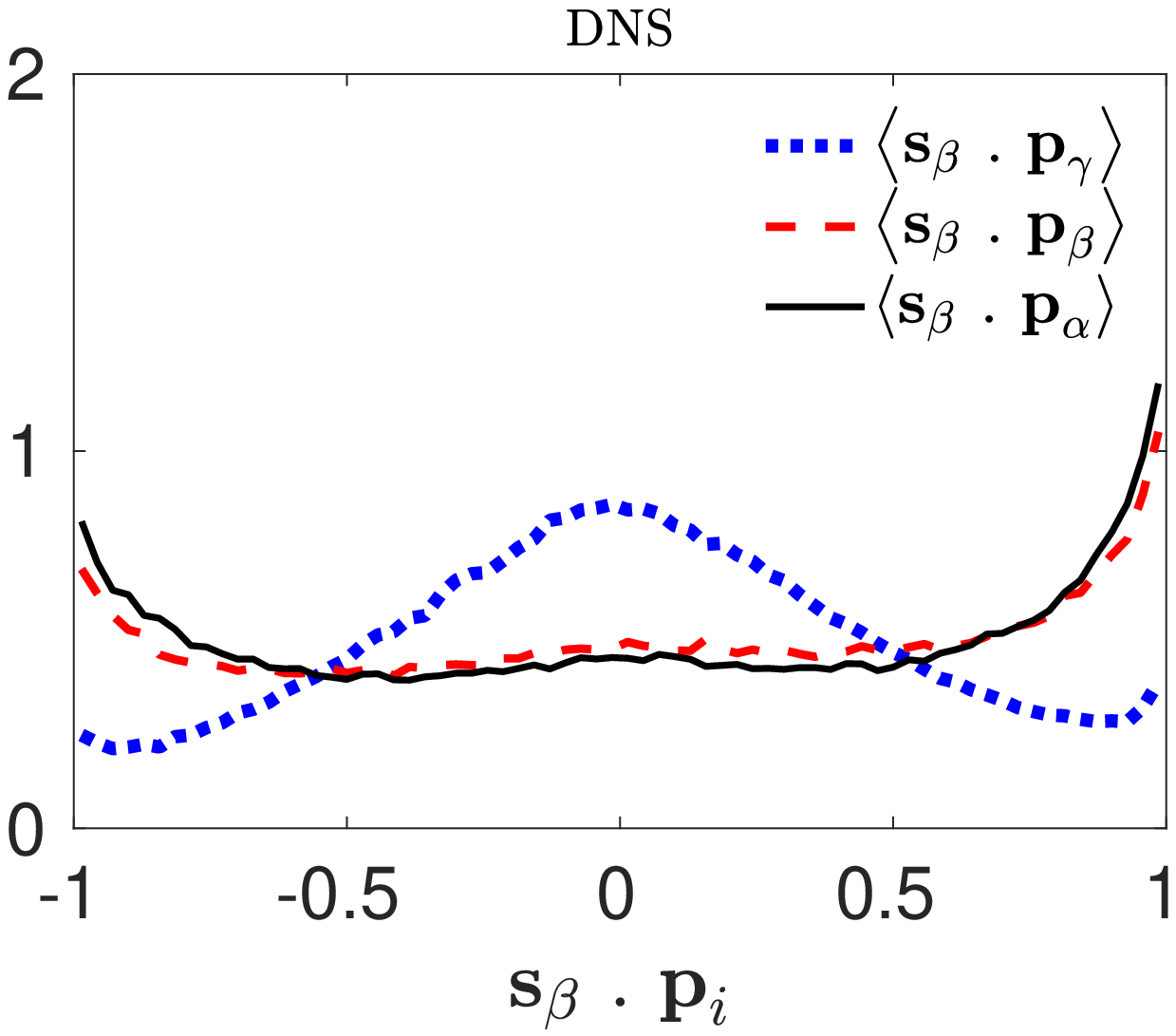}}
    \subfigure[]{
    \includegraphics[width=4.5cm]{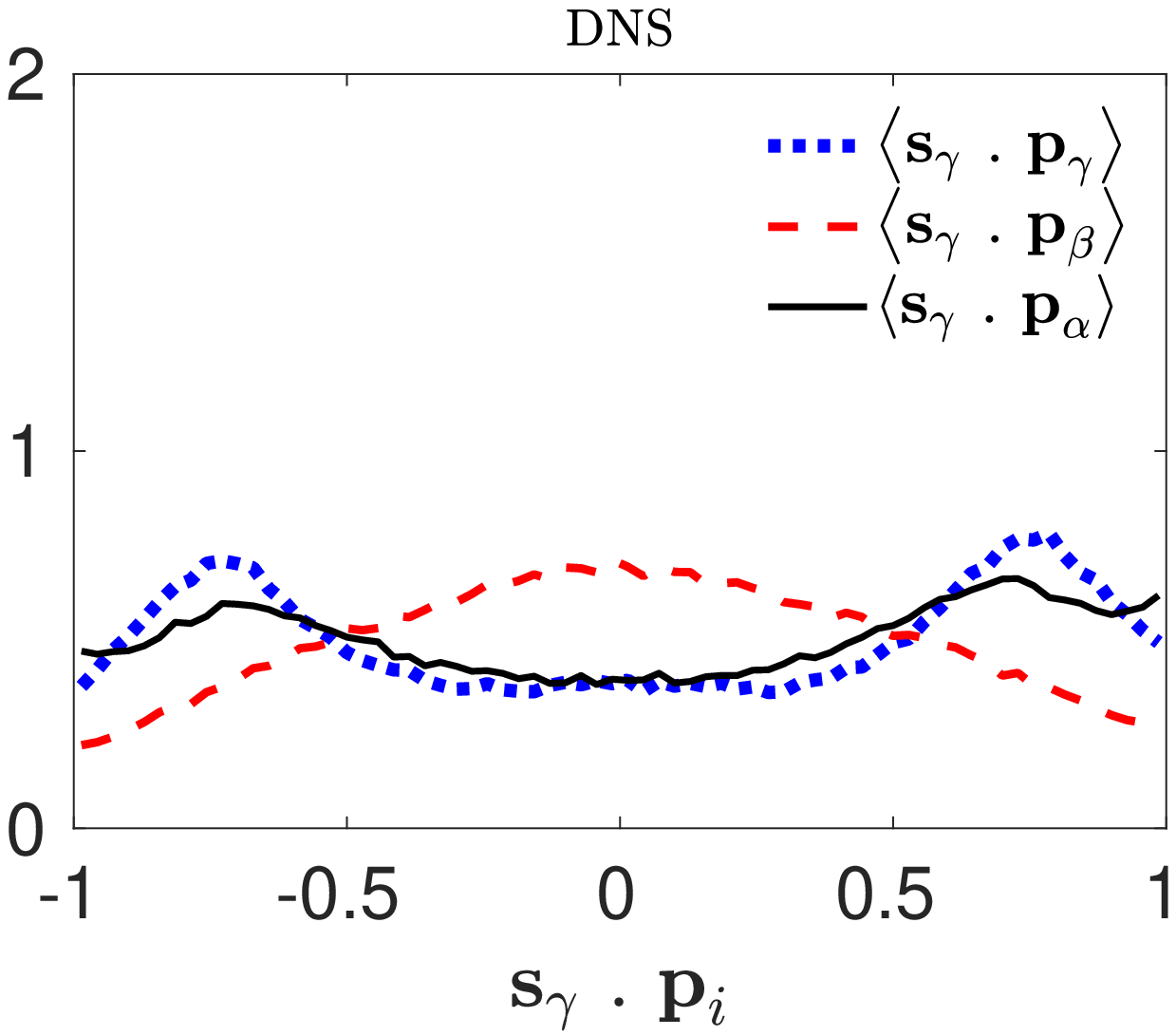}}
    \caption{Alignment of $\boldsymbol{\mathcal{P^{RFD}}}$-eigenvectors ($\boldsymbol{p_i}$) with $\boldsymbol{S}$-eigenvectors ($\boldsymbol{s_i}$). Here, $i$ (= $\alpha$, $\beta$ or $\gamma$) denotes the three eigenvectors corresponding to the three eigenvalues $\alpha>\beta>\gamma$}
    \label{fig:s_dot_p_rfdm}
    \end{figure}

    \item \textit{In strain-dominated regions, the eigenvectors of $\boldsymbol{\mathcal{P}^{RFD}}$ coincides with the strain-rate eigenvectors:} 
    \par
    As discussed above $\boldsymbol{\mathcal{P}^{RFD}}$ is always either negative-definite or positive-definite. Further, it is evident that if $\boldsymbol{A}$ is close to being symmetric (strain-dominant, $\boldsymbol{A}\approx\boldsymbol{S}$), the eigenvectors of $\boldsymbol{\mathcal{P}^{RFD}}$ will be approximately parallel or perpendicular to the eigenvectors of $\boldsymbol{S}$ itself. Hence, in strain-dominated regions $\boldsymbol{\mathcal{P}^{RFD}}$ is expected to show biased alignment towards the strain-rate eigenvectors which is nonphysical. In order to verify this claim, we show the alignment of the eigenvectors of $\boldsymbol{\mathcal{P}}$ and $\boldsymbol{\mathcal{P}^{RFD}}$ with strain-rate eigenvectors in Figure \ref{fig:s_dot_p_rfdm}. In Figure \ref{fig:s_dot_p_rfdm}(a,b,c) we show the PDF (probability distribution function) of the alignment of eigenvectors of $\boldsymbol{\mathcal{P}^{RFD}}$ with strain-rate eigenvectors and in Figure \ref{fig:s_dot_p_rfdm}(d,e,f) we show alignment of $\boldsymbol{\mathcal{P}}$-eigenvectors with $\boldsymbol{S}$-eigenvectors for comparison. It can be observed that for a large percentage of particles $\boldsymbol{\mathcal{P^{RFD}}}$-eigenvectors are either parallel or perpendicular to the $S$-eigenvectors (Figure \ref{fig:s_dot_p_rfdm}(a,b,c)). On the other hand, the eigenvectors of $\boldsymbol{\mathcal{P}}$ obtained from DNS show no such alignment tendencies as shown by $\boldsymbol{\mathcal{P^{RFD}}}$-eigenvectors. 

\end{enumerate}

\section{Using Neural networks to model pressure-Hessian}
\label{s:NN}
It is evident from the discussion in the previous section that the functional relationship between pressure-Hessian and local velocity gradient tensor, if any, is far too complex to be addressed by simple algebraic models. In general, the evolution of pressure-Hessian of individual fluid particles is expected to be governed by a large spectrum of flow quantities, their higher derivatives and their evolutionary history as well. Nevertheless, in this work, we intend to explore the maximum potential of local velocity gradients to describe the pressure-Hessian accurately. For this purpose, we employ deep neural networks. Given, a sufficiently large network and training-data, neural networks can potentially decipher the functional relationship (if any) existing between the quantities of interest. With this motivation, we resort to neural networks to provide a better mapping between pressure-Hessian and the velocity-gradient tensor.

\subsection{Neural network architecture}
\begin{figure}[bt]
\centering
\includegraphics[width=17cm]{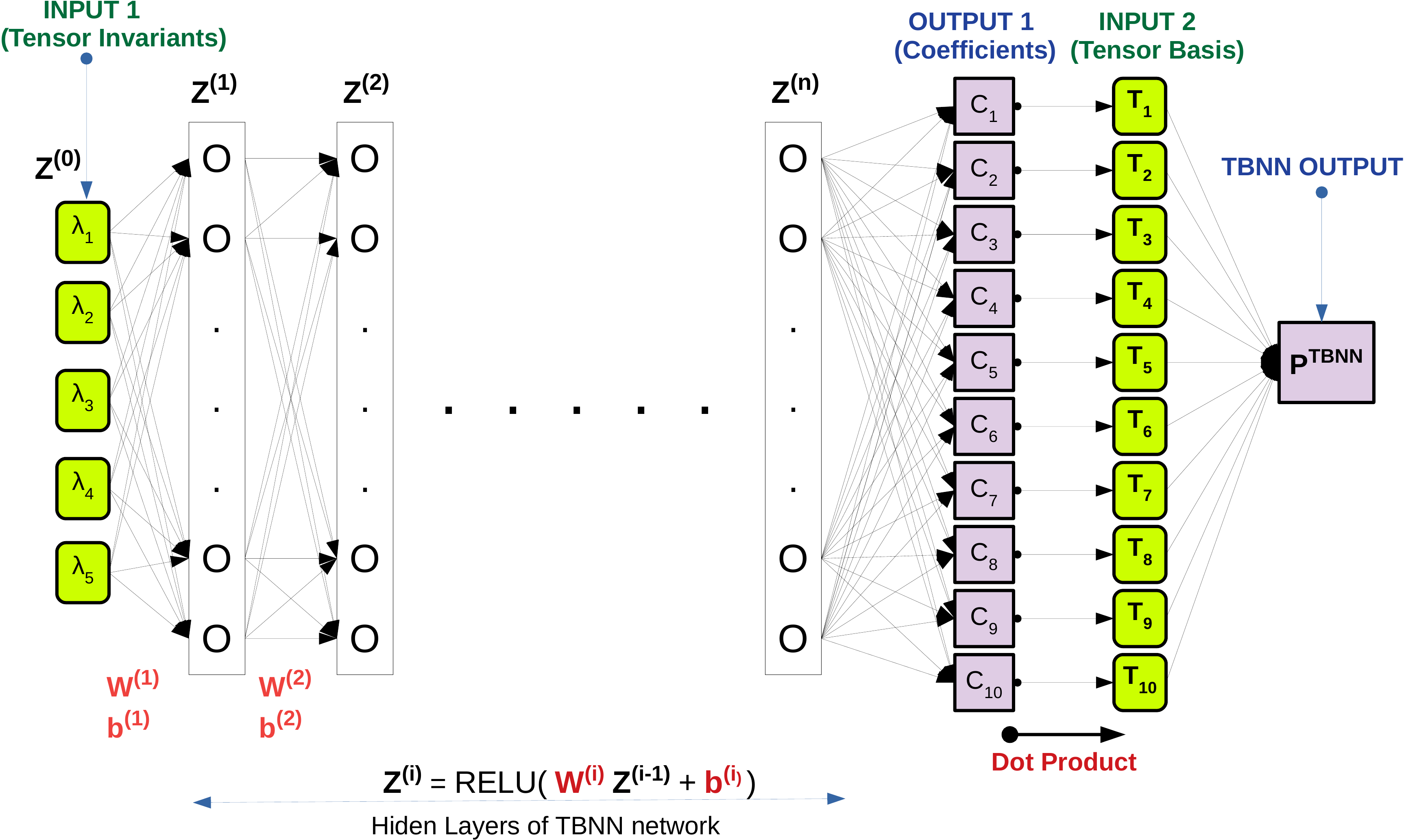}
\caption{Schematic of the TBNN network. $W^{(i)}$ and $b^{(i)}$ are the weight matrix and the bias vector of the $i^{th}$ layer. Both $W^{(i)}$ and $b^{(i)}$ are the learnable parameters of the neural network, which are optimized using the RMSprop optimizer \cite{RMSprop}.}
\label{fig:tbnn}
\end{figure}

In this work, we employ the tensor basis neural network (TBNN) developed by \citet{ling2016}. This architecture has been shown to be robust for mapping tensors. The TBNN increases the representation power of the neural network by embedding knowledge of Tensor basis ($\boldsymbol{T}^{i}$) and invariants ($\lambda^{i}$) in the network itself. The TBNN network takes advantage from the Caley-Hamilton theorem, which states that any function derived from a given tensor alone can be expressed as a linear combination of the integrity basis \cite[]{spencer1958} of the given tensors. The predictions made by the TBNN network are basically a linear combination of the integrity basis ($\boldsymbol{T}^{i}$) of the input tensors. Hence, the TBNN network explores the full spectrum of all the mappings that any input tensor can offer, by enforcing the output of the network to be a linear combination of its integrity basis. Further, the TBNN network has embedded rotational invariance, which ensures that the predictions made by the TBNN network are independent of the orientation of the coordinate system. If the input tensors are expressed in a rotated-coordinate system, the predicted output will also get rotated accordingly. Hence, the TBNN network predicts the same output tensor irrespective of the orientation of the coordinate system. Figure \ref{fig:tbnn}, presents a brief overview of the TBNN network.

For incompressible flow field, \citet{pope1975} derived the ten trace-free integrity basis ($\boldsymbol{T}^i$) and five independent invariants ($\lambda^i$) of the strain-rate ($\boldsymbol{S}$) and rotation rate ($\boldsymbol{R}$) tensors. 
These tensor basis and invariants are listed below:
\begin{align}
    \boldsymbol{T}^1 &= \boldsymbol{S}, \ &\boldsymbol{T}^2 &= \boldsymbol{SR}-\boldsymbol{RS},  \nonumber\\
    \boldsymbol{T}^3 &= \boldsymbol{S}^2-\frac{1}{3}\boldsymbol{I}\{\boldsymbol{S}^2\}, \ &\boldsymbol{T}^4 &= \boldsymbol{R}^2-\frac{1}{3}\boldsymbol{I}\{\boldsymbol{R}^2\},  \nonumber\\
    \boldsymbol{T}^5 &= \boldsymbol{RS}^2-\boldsymbol{S}^2\boldsymbol{R}, \ &\boldsymbol{T}^6 &= \boldsymbol{R}^2\boldsymbol{S}+\boldsymbol{SR}^2-\frac{2}{3}\boldsymbol{I}\{\boldsymbol{SR}^2\},  \nonumber\\
    \boldsymbol{T}^7 &= \boldsymbol{RSR}^2-\boldsymbol{R}^2\boldsymbol{SR}, \ &\boldsymbol{T}^8 &= \boldsymbol{SRS}^2-\boldsymbol{S}^2\boldsymbol{RS},  \nonumber\\
    \boldsymbol{T}^9 &= \boldsymbol{R}^2\boldsymbol{S}^2+\boldsymbol{S}^2\boldsymbol{R}^2-\frac{2}{3}\boldsymbol{I}\{\boldsymbol{S}^2\boldsymbol{R}^2\},   \ &\boldsymbol{T}^{10} &= \boldsymbol{RS}^2\boldsymbol{R}^2-\boldsymbol{R}^2\boldsymbol{S}^2\boldsymbol{R}; 
    \label{eq:basis}
\end{align}
\begin{align}
    \lambda^1 = \{\boldsymbol{S}^2\},  \hspace{0.65cm}  \lambda^2 = \{\boldsymbol{R}^2\},  \hspace{0.65cm}  \lambda^3 = \{\boldsymbol{S}^3\}, \hspace{0.65cm}  \lambda^4 = \{\boldsymbol{R}^2\boldsymbol{S}\}, \hspace{0.65cm}  \lambda^5 = \{\boldsymbol{R}^2\boldsymbol{S}^2\}. 
    \label{eq:invariants}
\end{align}
The symbol $\{\}$ represents the trace of the tensor.
A linear combination of these ten tensor basis ($\boldsymbol{T}^i$) can represent any trace-free tensor that is directly derived from $\boldsymbol{S}$ and $\boldsymbol{R}$. Since the exact expression for the trace of the pressure-Hessian is already known:
\begin{equation}
    \{\boldsymbol{\mathcal{P}}\} = -A_{ik}A_{ki}, 
\end{equation}
these trace-free integrity bases ($T_i$) can be readily used to model the trace-free part of the pressure-Hessian using the TBNN. We use the symbol $\boldsymbol{\mathcal{P}_{tf}}$ to denote the trace-free part of $\boldsymbol{\mathcal{P}}$. To find the relevant mapping between velocity gradient tensor and $\boldsymbol{\mathcal{P}_{tf}}$ the ten coefficients ($C^i$) corresponding to the ten integrity basis ($\boldsymbol{T}^i$) needs to be modelled. The five invariants ($\lambda^i$) of $\boldsymbol{S}$ and $\boldsymbol{R}$ forms the primary input of the TBNN. The output of the last layer of the network yields the ten coefficients $C^i$. A secondary input containing the ten tensor basis $\boldsymbol{T}^i$ (called tensor layer) is fed to the last layer of the network. Finally, a dot product between the coefficient layer and the tensor layer of the network makes the final output of the network, which can be expressed as:
\begin{equation}
    \boldsymbol{\mathcal{P}_{tf}^{TBNN}} = \sum_{i=1}^{10} C^i \boldsymbol{T}^i.
\end{equation}

The cost function of the network can be expressed as:
\begin{equation}
    J = \frac{1}{2m}\sum_{j=1}^m \left[\left|\left|\left(\boldsymbol{\mathcal{P}^{TBNN}_{tf}} - \boldsymbol{\mathcal{P}_{tf}}\right)_j\right|\right|_{F}\right]^2,
\end{equation}
where $m$ is the number of training examples required to train the TBNN and the symbol $|| \ ||_F$ represents the Frobenius norm.
\subsection{Training of the neural network}
The employed tensor basis neural network (TBNN) model is trained using data from an isotropic incompressible flow field at Reynolds number of 433. This data is taken from the John Hopkins University's Turbulence database \cite[]{JHUTD_1, JHUTD_2} available online at \url{http://turbulence.pha.jhu.edu/}. The opensource library Keras \cite[]{chollet2015keras} with TensorFlow backend is used for training the TBNN model. The velocity gradient tensor and pressure-Hessian information are extracted from the database at a particular time instant. A total number of 262,144 unique data-points are extracted from the flow field. Out of these 262,144 data points, 236,544 points are used for training the network, while the remaining 25,600 data-points are reserved for the cross-validation of the predicted solution. The training data is randomly distributed into 924 mini-batches of 256 data-points each at the beginning of every epoch. Since one epoch is one complete pass through the training dataset overall mini-batches. Hence, one epoch accounts for 924 iterations of the training cycle.
The velocity gradient tensor was non-dimensionalized with the mean value of the Frobenius norm of the whole sample of 262,144 data-points. No, further normalization was used for the derived tensor-basis ($\boldsymbol{T}^i$) and invariants ($\lambda^i$).
\begin{figure}[bt]
\centering
\includegraphics[width=8cm]{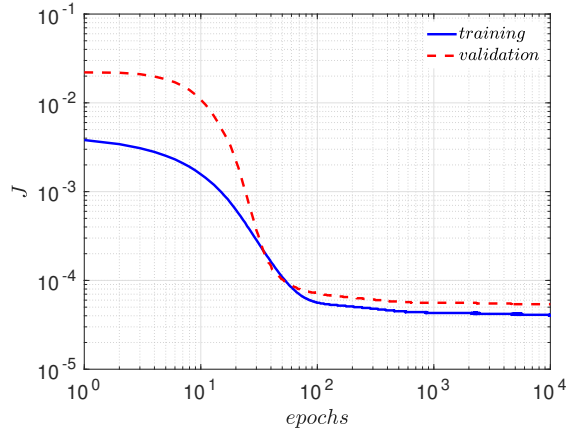}
\caption{Decay of cost function during training for TBNN. Mini-batch size=256, 1 epoch = 924 iterations of the optimizer.}
\label{fig:cost}
\end{figure}

A deep network with 11 hidden layers and a combination of 50, 150, 150, 150, 150, 300, 300, 150, 150, 150, 100 in the consecutive hidden layers was found to yield the best performance of all the combinations that were tested. We use the Glorot normal initialization \cite[]{glorot2010} for weight matrices and RELU (rectified linear unit) non-linear activation function for the hidden layers. The RMSprop optimizer \cite[]{RMSprop}, with a learning rate of $1.0e-6$ was used to train the network. The training was stopped when the value of cost function $J$ became stagnant. The minimum value of training cost and cross-validation cost recorded while training was $4.1e$-$4$ and $5.4e$-$4$ respectively. In Figure \ref{fig:cost}, we show the training and cross-validation cost as a function of a number of training epochs. The cross-validation cost didn't show any significant rise during the training process. A low dropout rate of 10$\%$ was used to facilitate ensemble learning in the network. There was no gain in model performance with further increase in data-size and network depth. 

\subsection{Testing of the trained network}
\begin{figure}[bt]
    \centering
    \subfigure[]{
    \includegraphics[width=4.5cm]{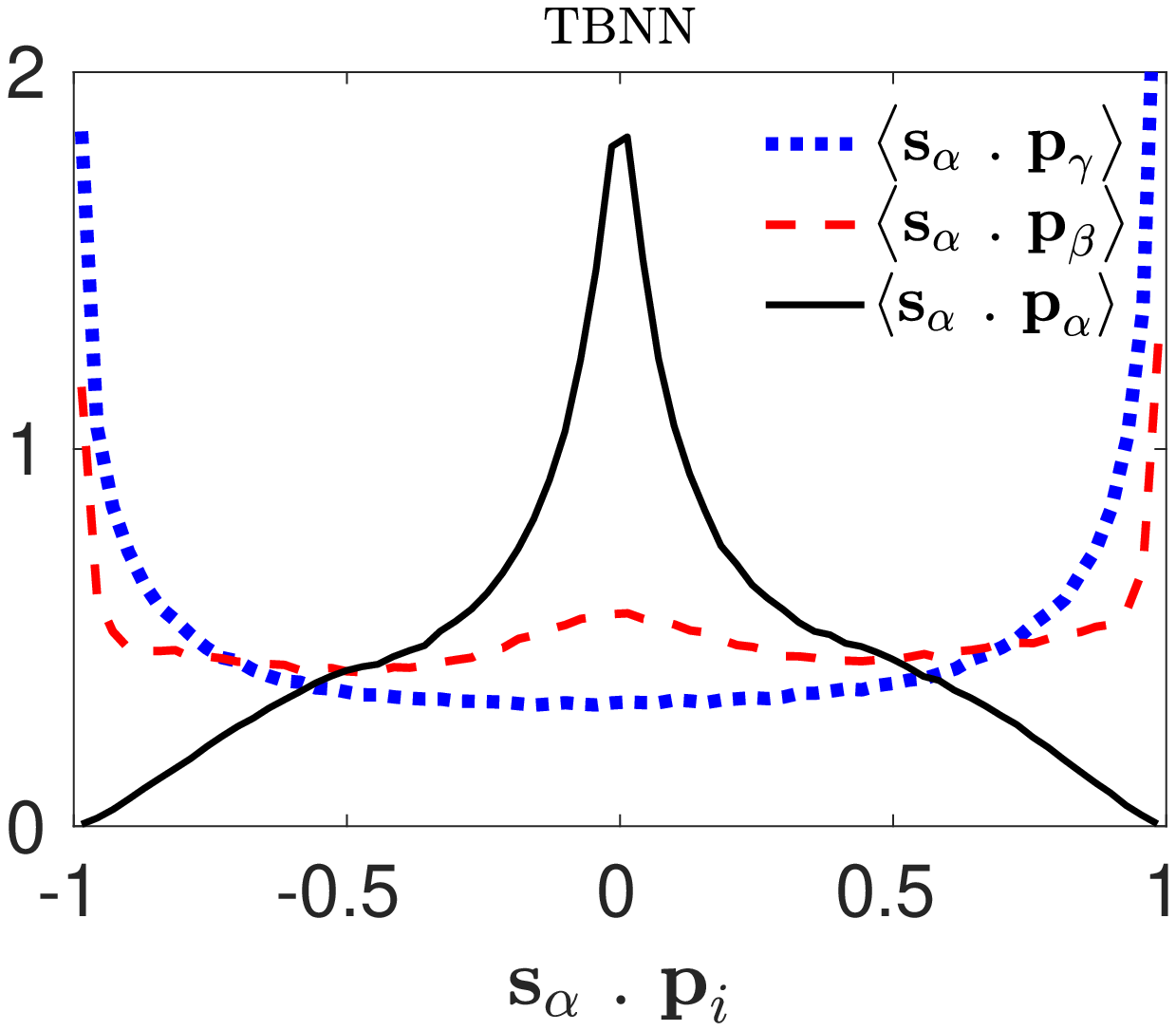}}
    \subfigure[]{
    \includegraphics[width=4.5cm]{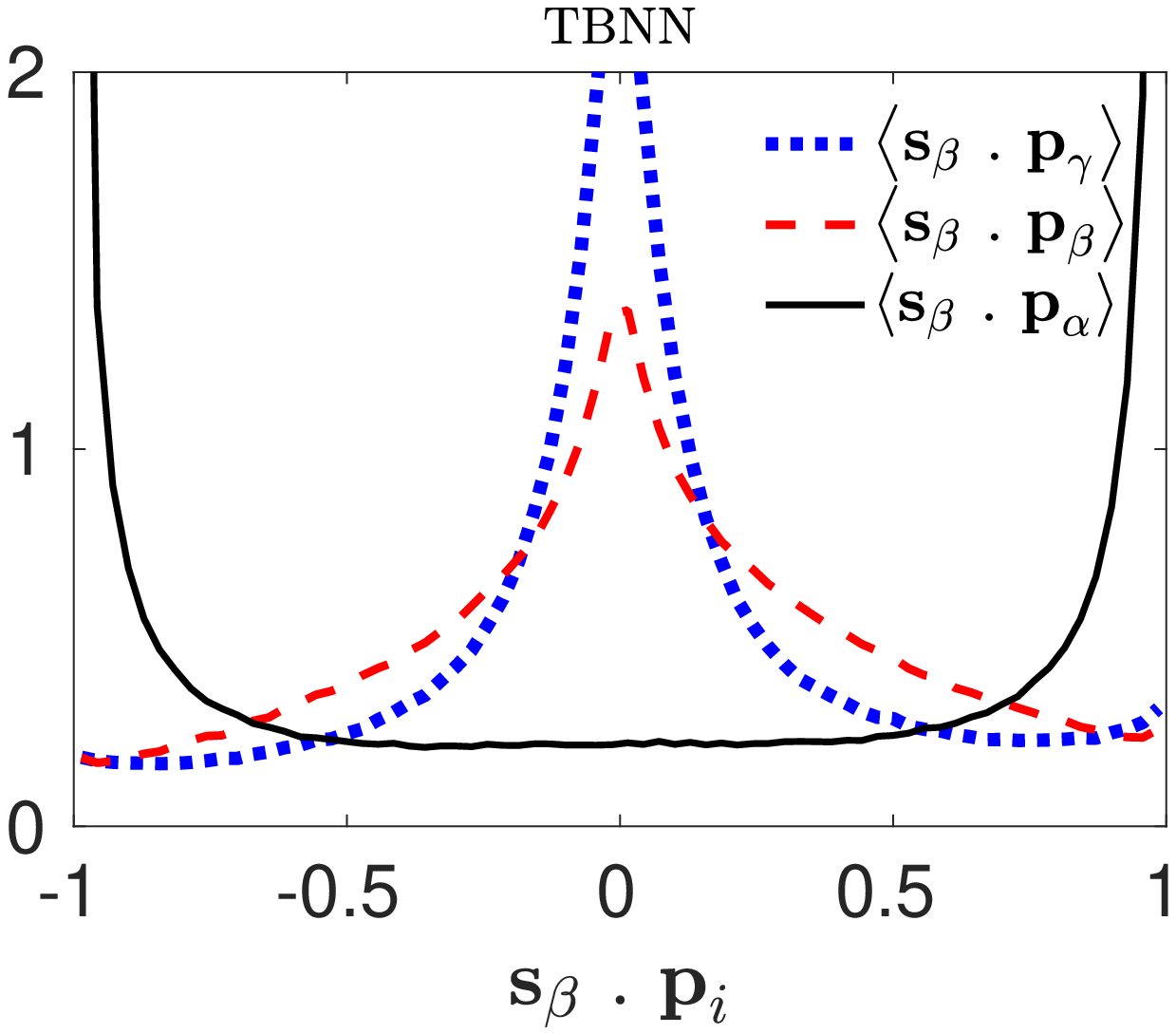}}
    \subfigure[]{
    \includegraphics[width=4.5cm]{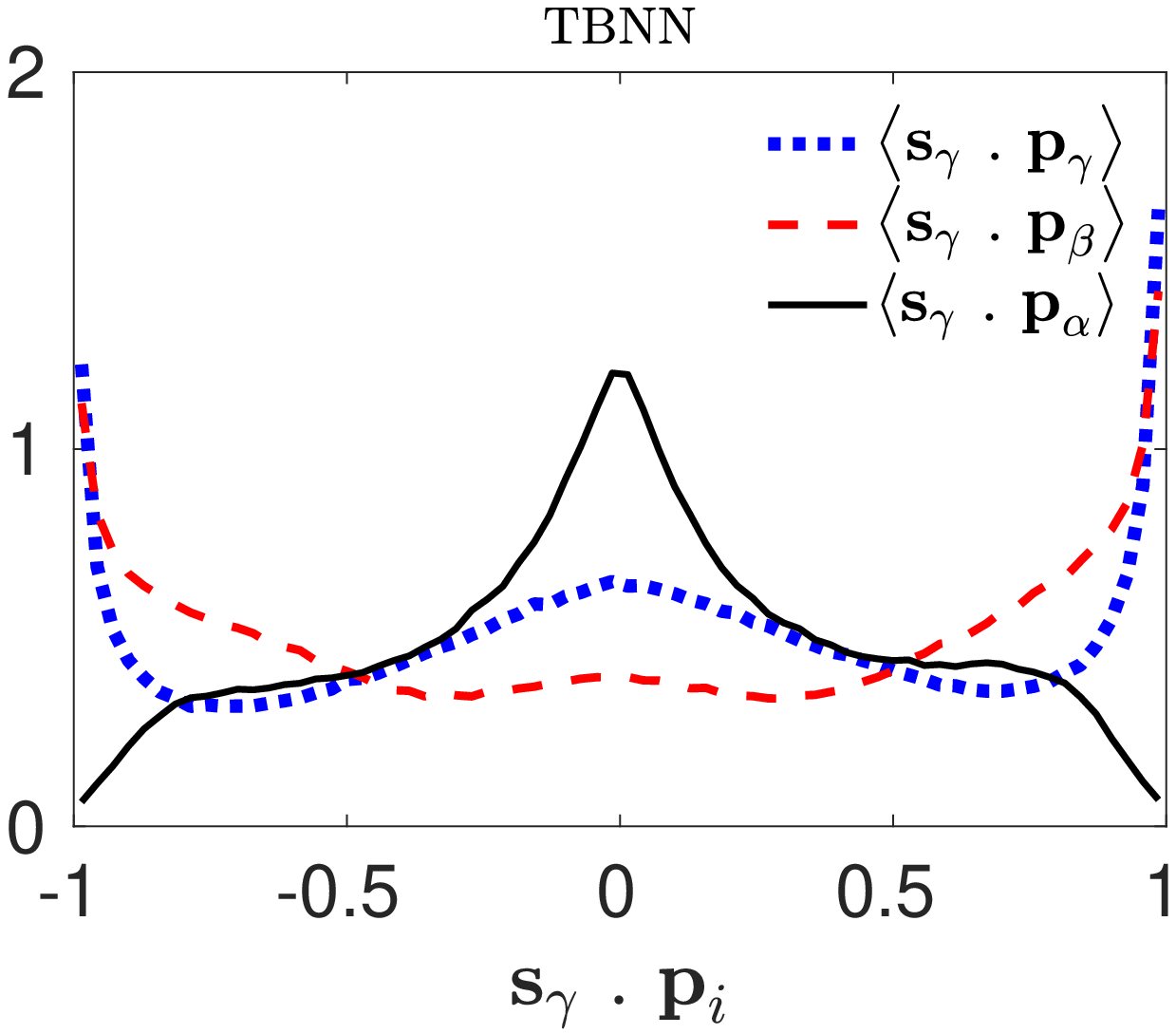}}
    \subfigure[]{
    \includegraphics[width=4.5cm]{s_dot_p_1_dns.eps}}
    \subfigure[]{
    \includegraphics[width=4.5cm]{s_dot_p_2_dns.eps}}
    \subfigure[]{
    \includegraphics[width=4.5cm]{s_dot_p_3_dns.eps}}
    \caption{Alignment of $\boldsymbol{\mathcal{P^{TBNN}}}$-eigenvectors ($\boldsymbol{p_i}$) with $\boldsymbol{S}$-eigenvectors ($\boldsymbol{s_i}$). Here, $i$ (= $\alpha$, $\beta$ or $\gamma$) denotes the three eigenvectors corresponding to the three eigenvalues $\alpha>\beta>\gamma$. (JHTD isotropic turbulence testing dataset, Reynolds number 433 \cite[]{JHUTD_1, JHUTD_2})}
    \label{fig:s_dot_p_tbnn}
\end{figure}

The primary testing of the trained TBNN model was performed on a separate testing dataset (other than training and validation data) of isotropic turbulence (JHTD \cite[]{JHUTD_1, JHUTD_2})
The relative Frobenius$-$norm error of the pressure-Hessian obtained from the trained model on the testing dataset was found to be $0.6491$. On the same dataset, an error of $0.7764$ was obtained by the RFDM model. Hence, in terms of element-wise mean squared error comparison, the accuracy of the trained TBNN model is comparable to the existing RFDM model. However, just the element-wise comparison is not a wise comparison metric for comparing tensorial quantities. We have earlier (in figure \ref{fig:s_dot_p_rfdm}) seen that the RFD model fails to capture the alignment statistics with the strain-rate tensor. In figure \ref{fig:s_dot_p_tbnn}, we present the alignment of the pressure-Hessian eigenvectors predicted by the TBNN (figure \ref{fig:s_dot_p_tbnn}(a,b,c)) with the strain-rate eigenvectors compared against that obtained from DNS (figure \ref{fig:s_dot_p_tbnn}(d,e,f)). We observe that although the alignment statistics (figure \ref{fig:s_dot_p_tbnn}(a,b,c)) have improved as compared to the RFD model results (figure \ref{fig:s_dot_p_rfdm}(a,b,c)), the obtained statistics are still far-off from that obtained from DNS.  

\section{Modified neural network architecture}
\label{s:NN_mod}
\begin{figure}[bt]
\centering
\includegraphics[width=8cm]{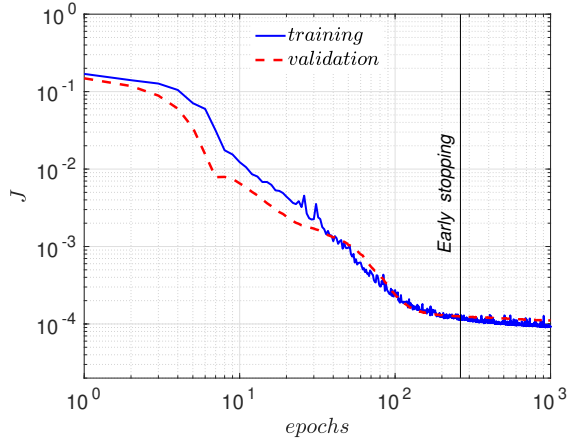}
\caption{Decay of cost function during training for the modified TBNN. Mini-batch size=256, 1 epoch = 924 iterations of the optimizer.}
\label{fig:cost_mod}
\end{figure}
\begin{figure}[h]
    \centering
    \subfigure[]{
    \includegraphics[width=4.5cm]{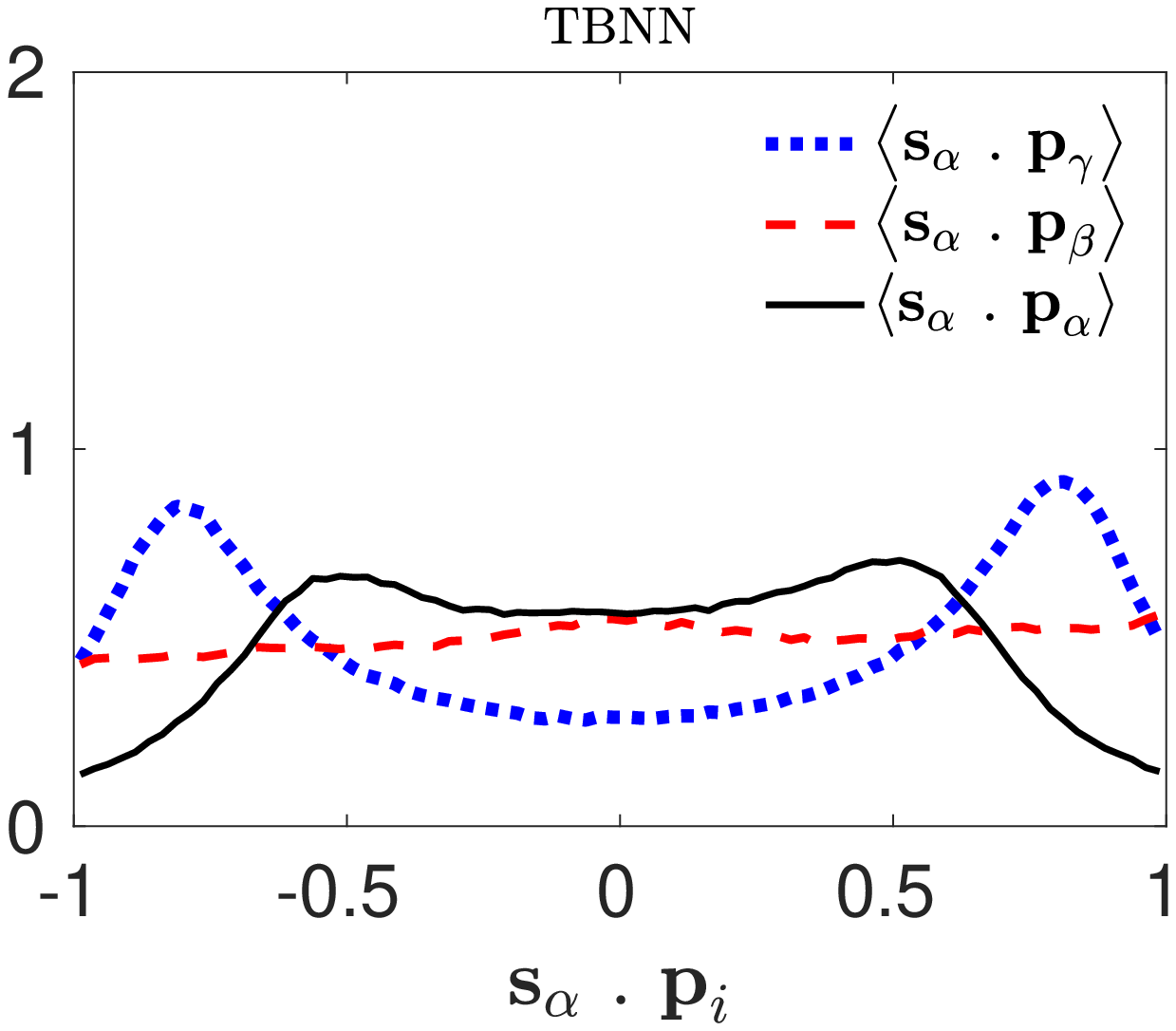}}
    \subfigure[]{
    \includegraphics[width=4.5cm]{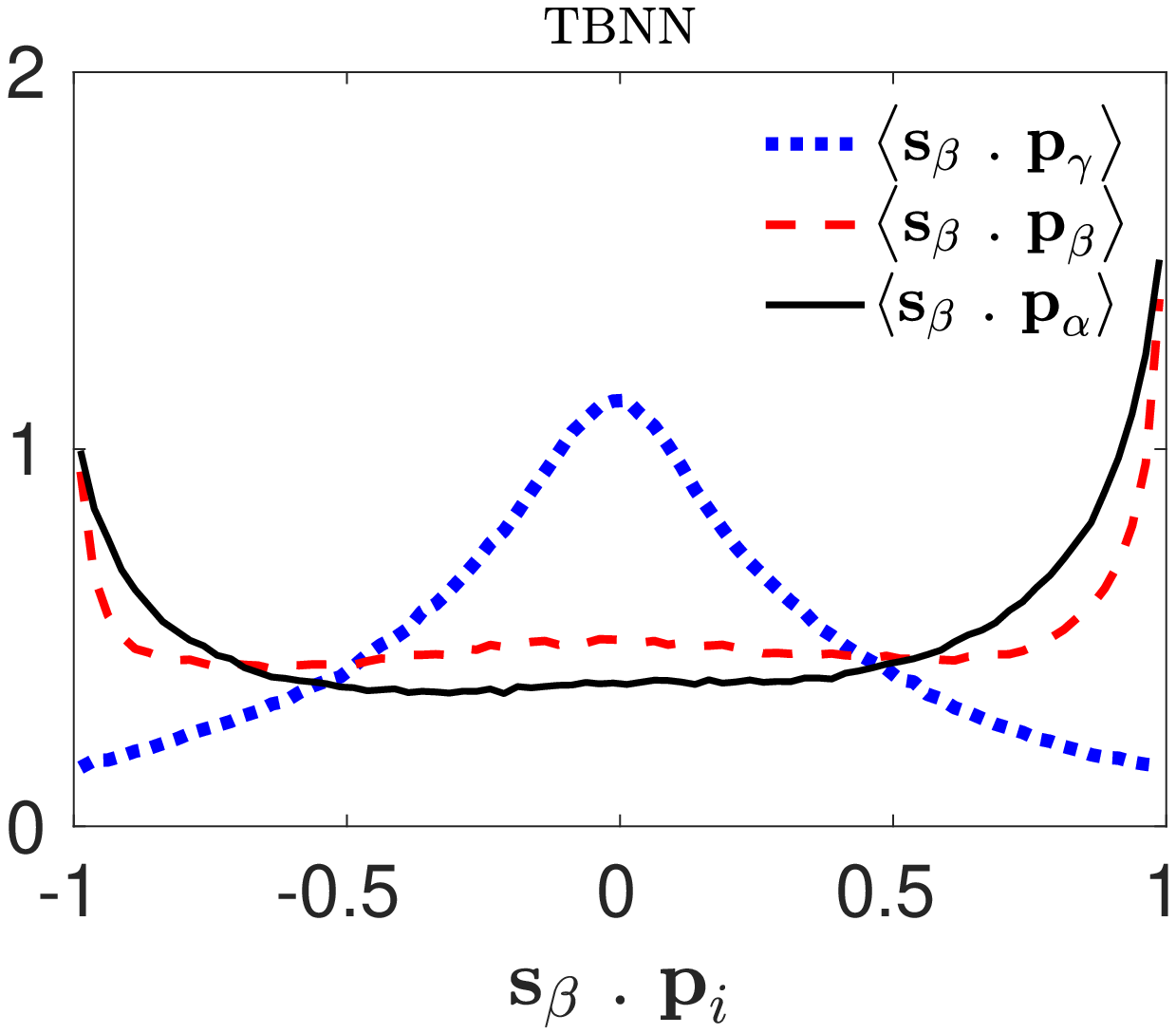}}
    \subfigure[]{
    \includegraphics[width=4.5cm]{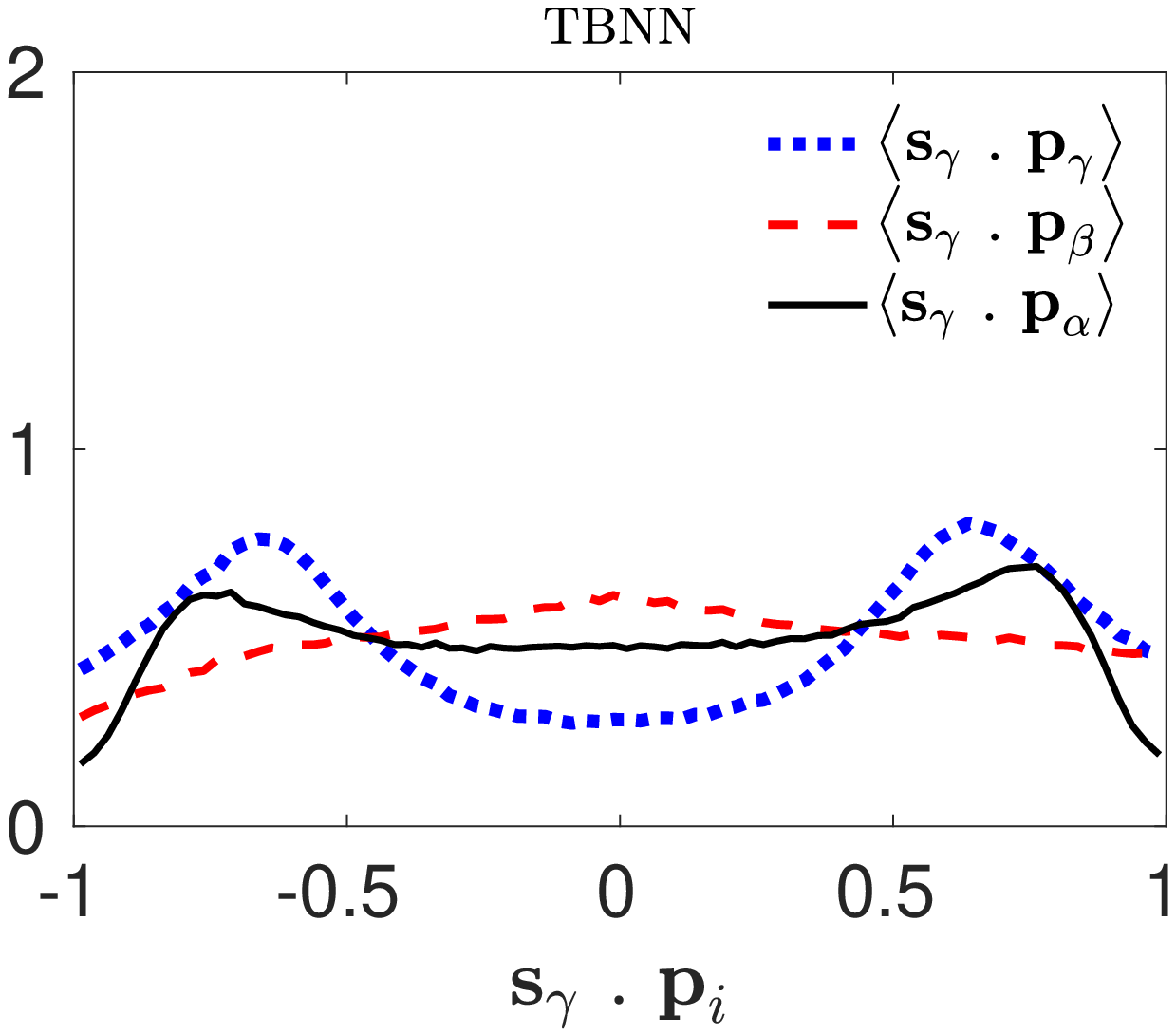}}
    \subfigure[]{
    \includegraphics[width=4.5cm]{s_dot_p_1_dns.eps}}
    \subfigure[]{
    \includegraphics[width=4.5cm]{s_dot_p_2_dns.eps}}
    \subfigure[]{
    \includegraphics[width=4.5cm]{s_dot_p_3_dns.eps}}
    \caption{Alignment of $\boldsymbol{\mathcal{P^{TBNN}}}$-eigenvectors ($\boldsymbol{p_i}$) obtained from modified TBNN with $\boldsymbol{S}$-eigenvectors ($\boldsymbol{s_i}$). Here, $i$ (= $\alpha$, $\beta$ or $\gamma$) denotes the three eigenvectors corresponding to the three eigenvalues $\alpha>\beta>\gamma$. (JHTD isotropic turbulence testing dataset \cite[]{JHUTD_1, JHUTD_2}, Reynolds number 433 \cite[]{JHUTD_1, JHUTD_2})}.
    \label{fig:s_dot_p_tbnn_mod}
\end{figure}

\begin{figure}[h]
    \centering
    \subfigure[]{
    \includegraphics[width=4.5cm]{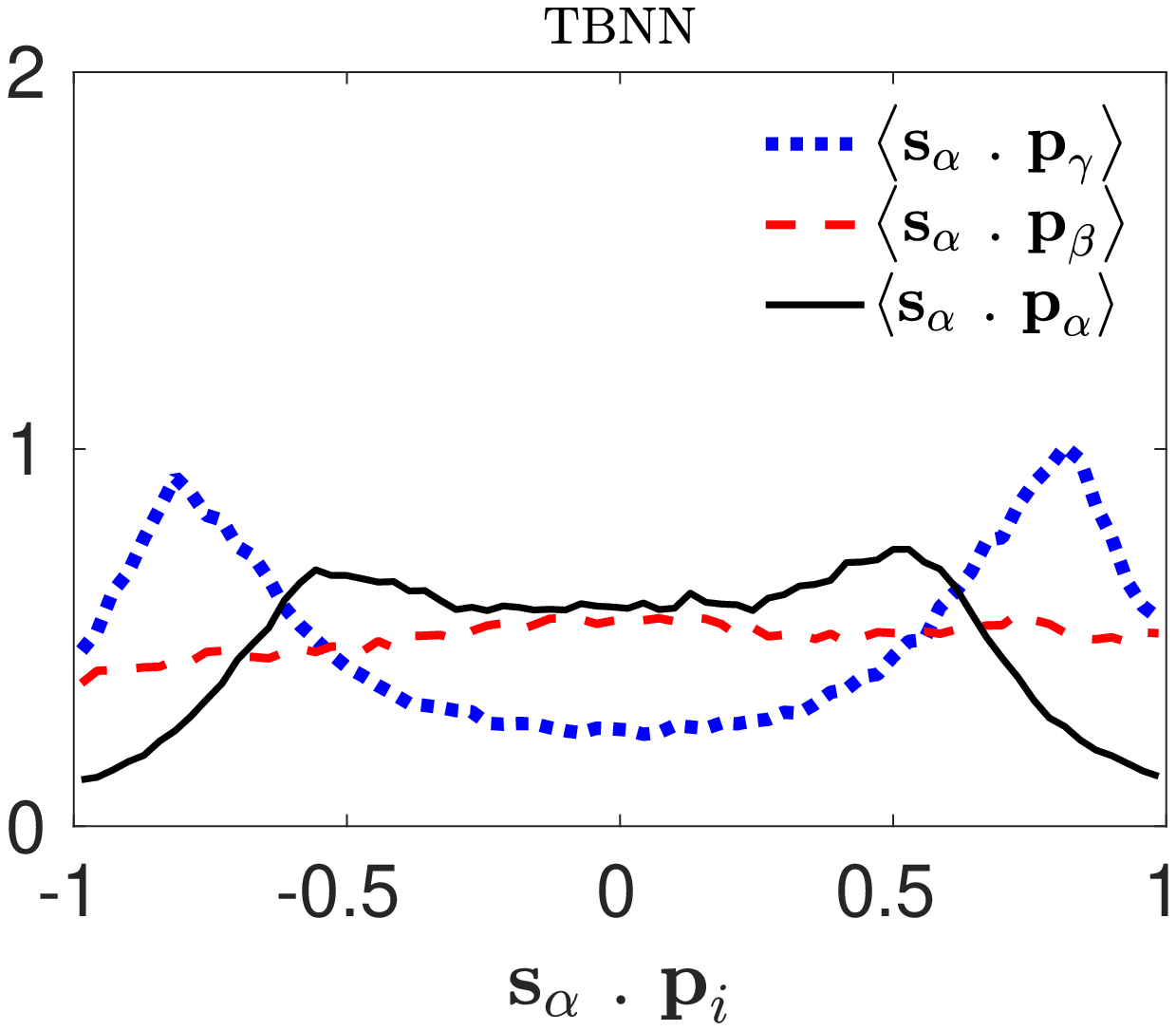}}
    \subfigure[]{
    \includegraphics[width=4.5cm]{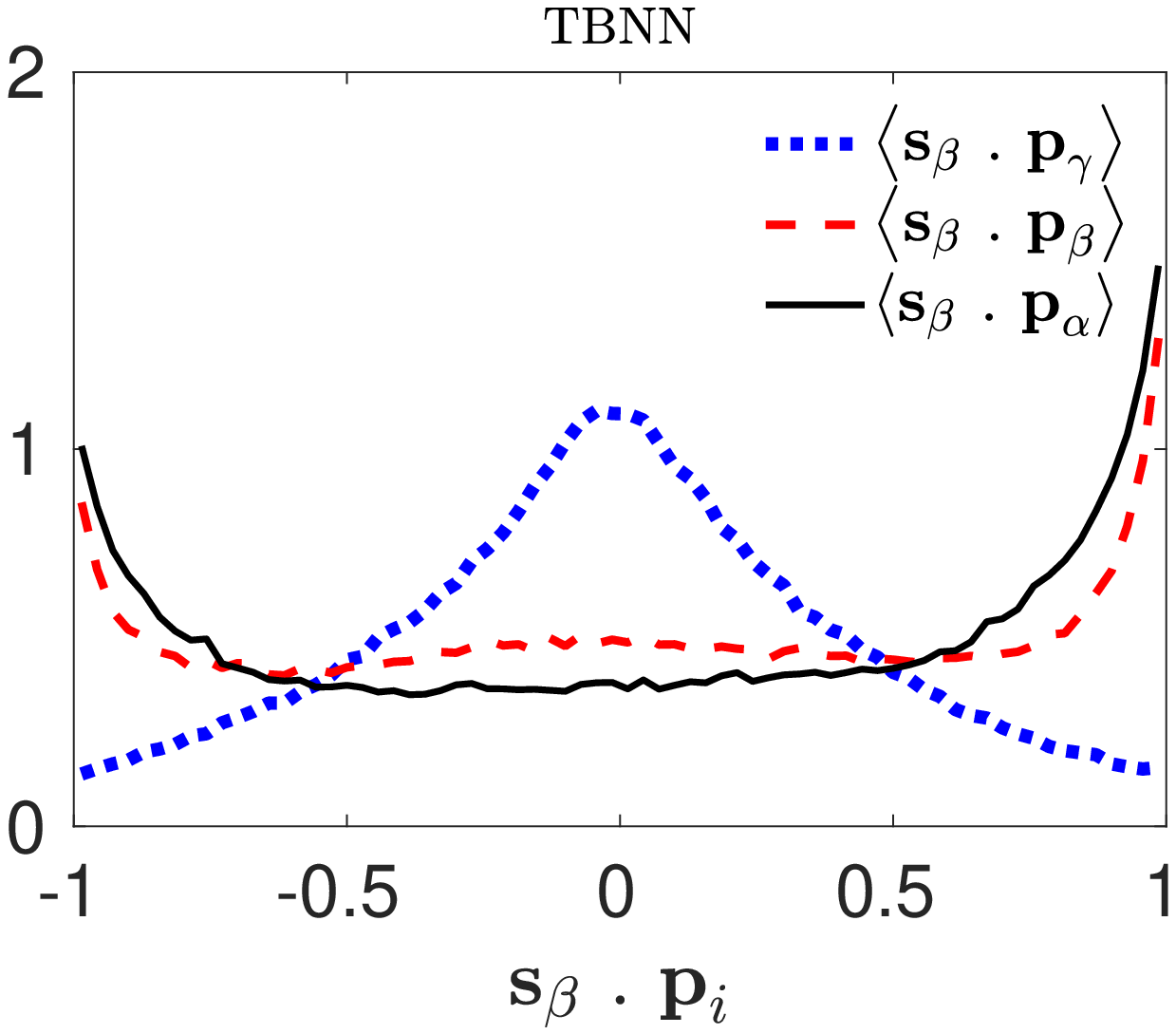}}
    \subfigure[]{
    \includegraphics[width=4.5cm]{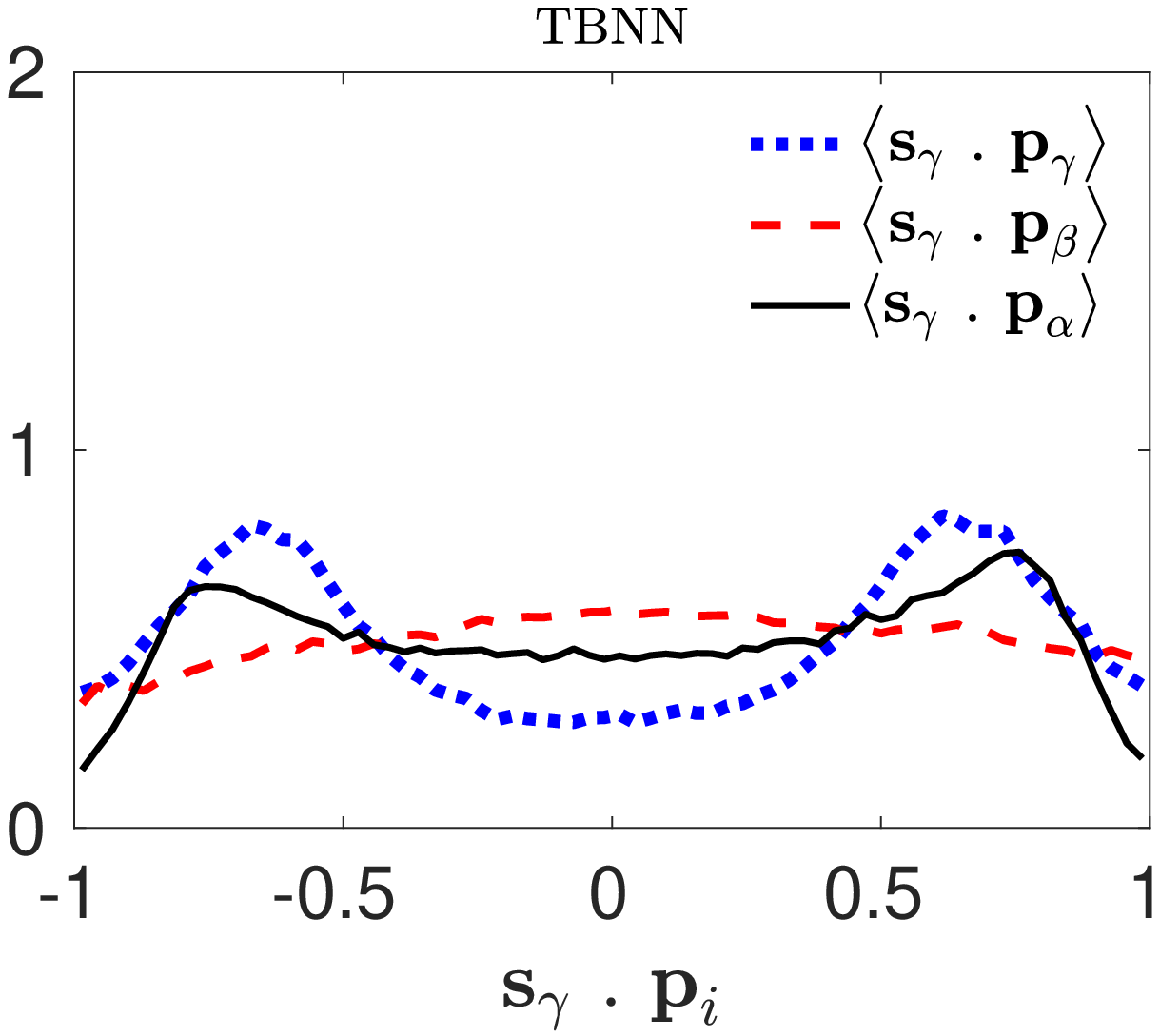}}
    \caption{Alignment of $\boldsymbol{\mathcal{P^{TBNN}}}$-eigenvectors ($\boldsymbol{p_i}$) obtained from modified TBNN with $\boldsymbol{S}$-eigenvectors ($\boldsymbol{s_i}$). Here, $i$ (= $\alpha$, $\beta$ or $\gamma$) denotes the three eigenvectors corresponding to the three eigenvalues $\alpha>\beta>\gamma$. (UP Madrid isotropic turbulence testing dataset \cite{cardesa2017}, Reynolds number 315)}.
    \label{fig:s_dot_p_tbnn_mod_madrid}
\end{figure}

We have observed that TBNN is unable to capture the alignment statistics of the pressure-Hessian tensor. It implies that assuming the pressure-Hessian to lie on the tensor basis of the strain-rate and rotation-rate tensors is not an appropriate modelling assumption. Constraining the network to obey tensor invariance properties restricts us to the use only global normalization of the input tensors. However, the velocity gradients in a turbulent flow field are known to be highly intermittent. Hence, global normalization of the input tensors might not be an effective strategy for such highly intermittent tensors. The learning of important feature mappings by a neural network relies heavily on effective normalization strategies. At this juncture, we performed several experiments on the TBNN by choosing various normalization strategies which allow TBNN to deviate from its tensor invariance characteristics. We found out through hit-and-trial that normalizing the tensor basis such that all its elements are scaled between [0, 1], yields tremendous improvement in network output. Two matrices $\bold{F}^{(i)}$ and $\bold{G}^{(i)}$ are used to scale the tensor basis:
\begin{equation}
{F^{(i)}}_{pq} = max \left({T^{(i)}}_{pq} \right)
\end{equation}
\begin{equation}
{G^{(i)}}_{pq} = min \left({T^{(i)}}_{pq} \right)
\end{equation}

Using $\bold{F}^{(i)}$ and $\bold{G}^{(i)}$ the tensor basis can be appropriately scaled using the following relationship:
\begin{equation}
    \boldsymbol{T}^{(i)'} = (\boldsymbol{T}^{(i)} - \boldsymbol{G}^{(i)}) \oslash (\boldsymbol{F}^{(i)} - \boldsymbol{G}^{(i)}),
\end{equation}
where, symbol $\oslash$ represents the Hadamard division between the two tensor.
With this normalization, the network loses most of the properties of the original TBNN. However, it leads to significant improvements in alignment statistics of the predicted output. 

We employ the modified network with the same settings (viz. the number of hidden layers, neurons per layer, activation function, learning rate, etc.) as used with the original TBNN network. In figure \ref{fig:cost_mod}, we show the learning curve obtained while training the modified TBNN. We use an early stopping criterion while training the network, at the point when the validation-loss curve becomes almost flat (no further decline with increasing epochs). 

\subsection{Testing modified TBNN for isotropic turbulence flow}
\label{ss:tbnn_iso}
In figure \ref{fig:s_dot_p_tbnn_mod} we show the alignment statistics obtained on the testing dataset with the modified TBNN on isotropic turbulence testing dataset (JHTD \cite[]{JHUTD_1, JHUTD_2}). We observe that modified TBNN predictions (figure \ref{fig:s_dot_p_tbnn_mod}(a,b,c)) demonstrate excellent alignment statistics as compared to that obtained from DNS (figure \ref{fig:s_dot_p_tbnn_mod}(d,e,f)). Although this testing dataset is extracted at different grid locations other than the training dataset, it still has the same Reynolds number as the training dataset (433). To make a better judgement of the generalization of the learnt pressure-Hessian mapping, we scrutinize the performance of the trained model for isotropic turbulence dataset at a different Reynolds number of 315. This dataset is extracted from the UP Madrid turbulence dataset \cite[]{cardesa2017}). In figure \ref{fig:s_dot_p_tbnn_mod_madrid} we plot the alignment statistics obtained from the learnt modified TBNN model for this dataset (at Renolds number of 315). We find that similar statistics are retrieved at Renolds number of 315 as well (Figure \ref{fig:s_dot_p_tbnn_mod_madrid}). Hence, we can conclude that the trained modified TBNN has learnt key physical features that can be generalized for an isotropic turbulent flow independent of its Reynolds number.

\subsection{Testing modified TBNN for turbulent channel flow}
\begin{figure}[h]
    \centering
    \subfigure[]{
    \includegraphics[width=4.5cm]{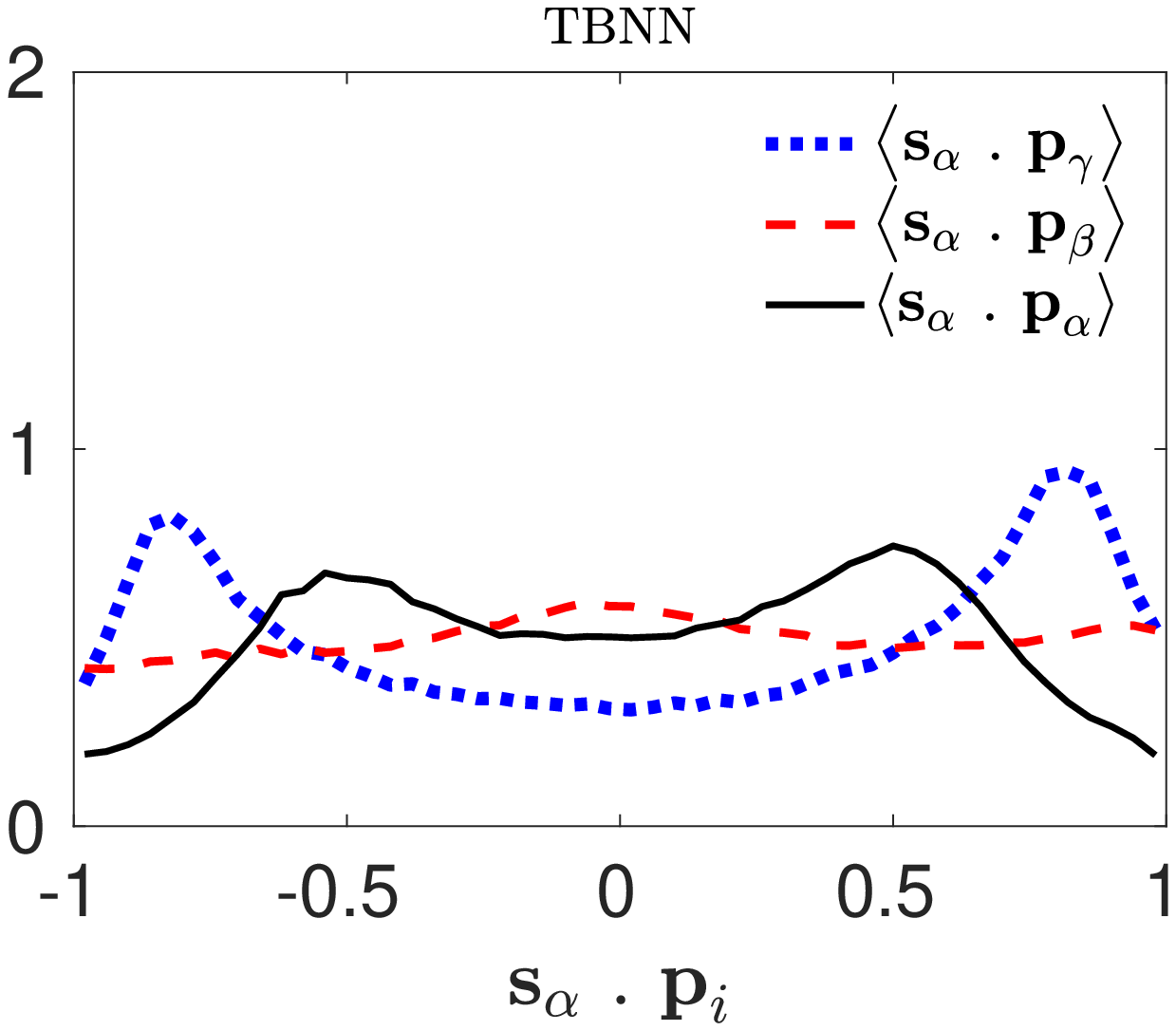}}
    \subfigure[]{
    \includegraphics[width=4.5cm]{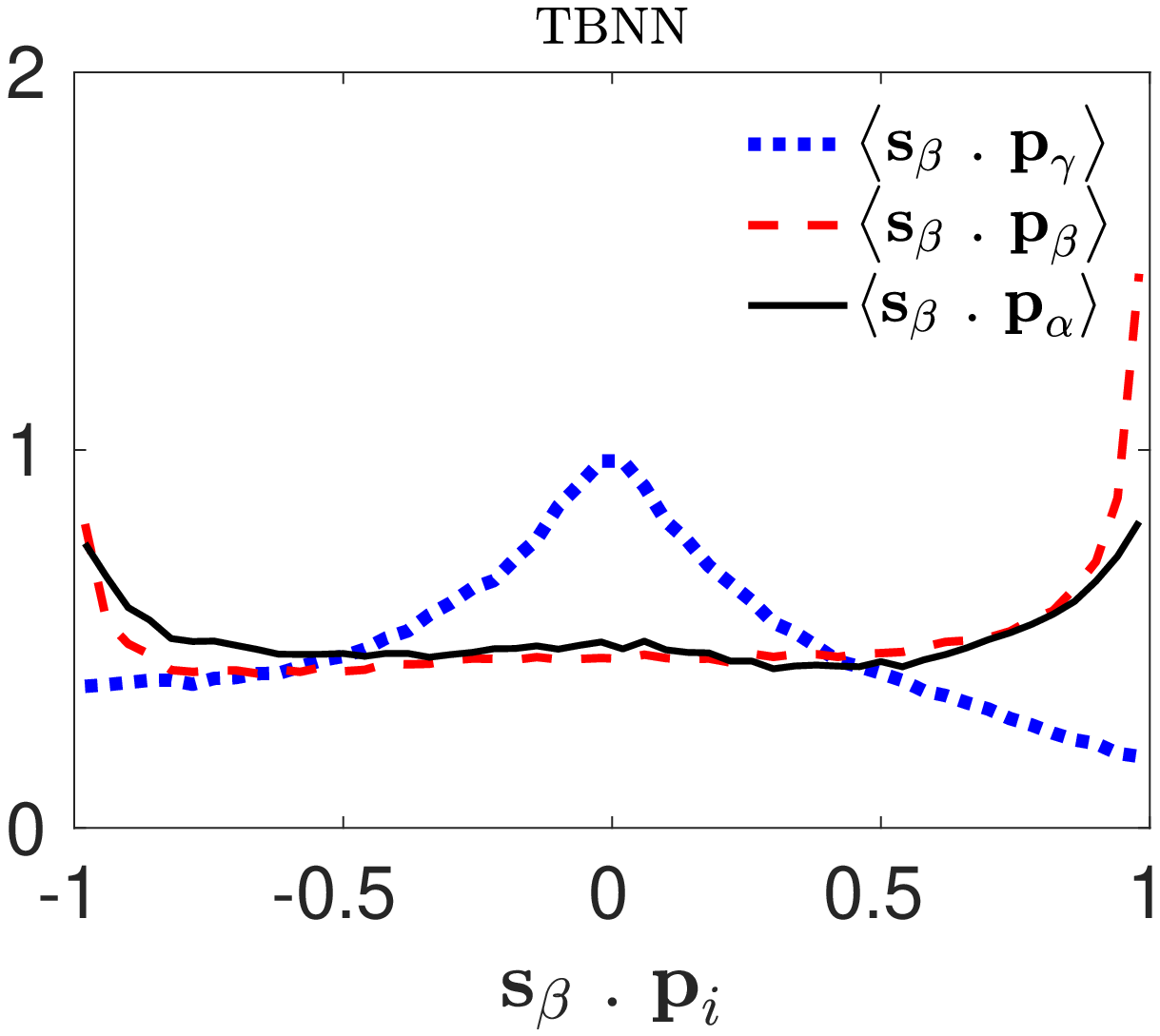}}
    \subfigure[]{
    \includegraphics[width=4.5cm]{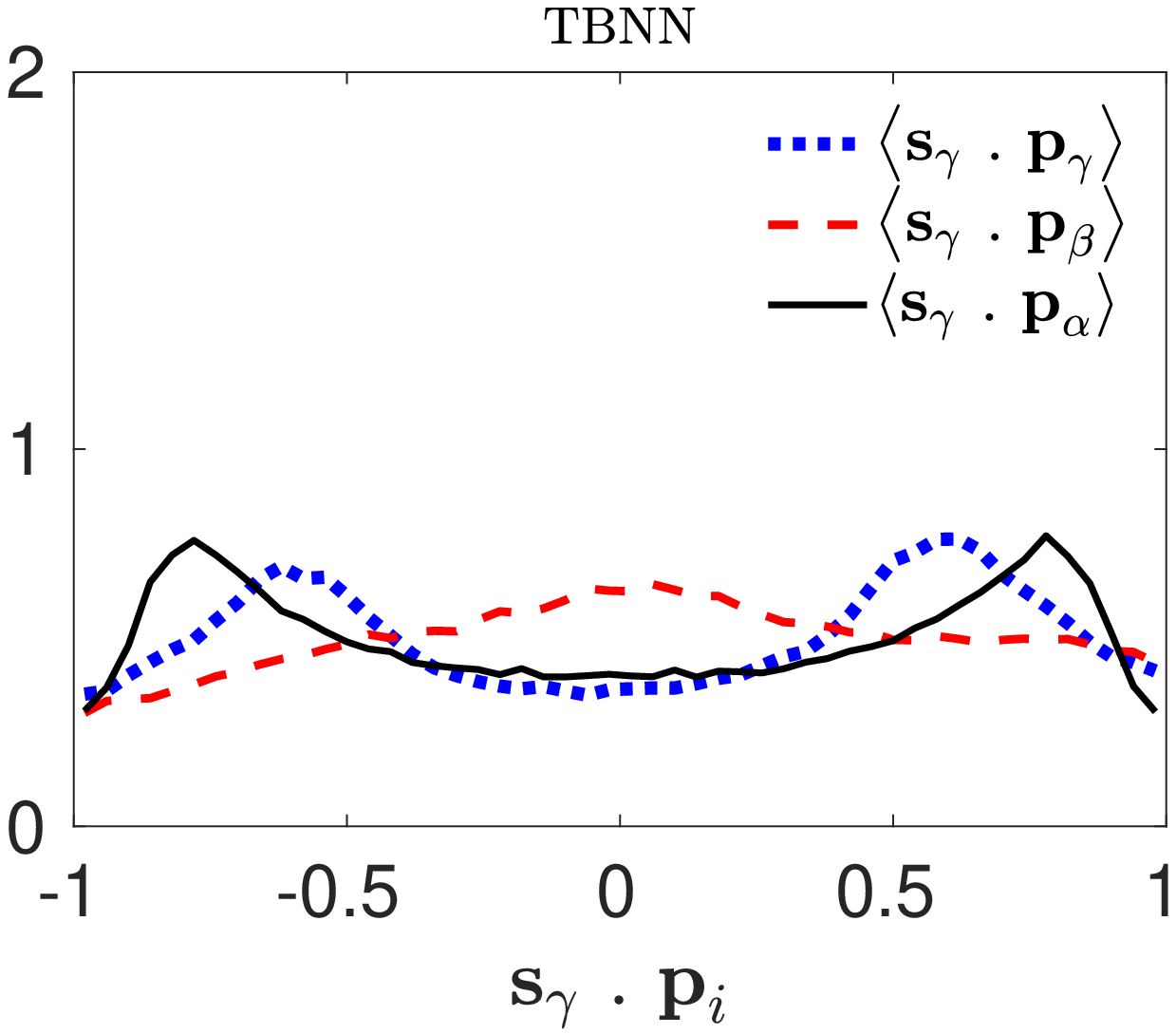}}
    \subfigure[]{
    \includegraphics[width=4.5cm]{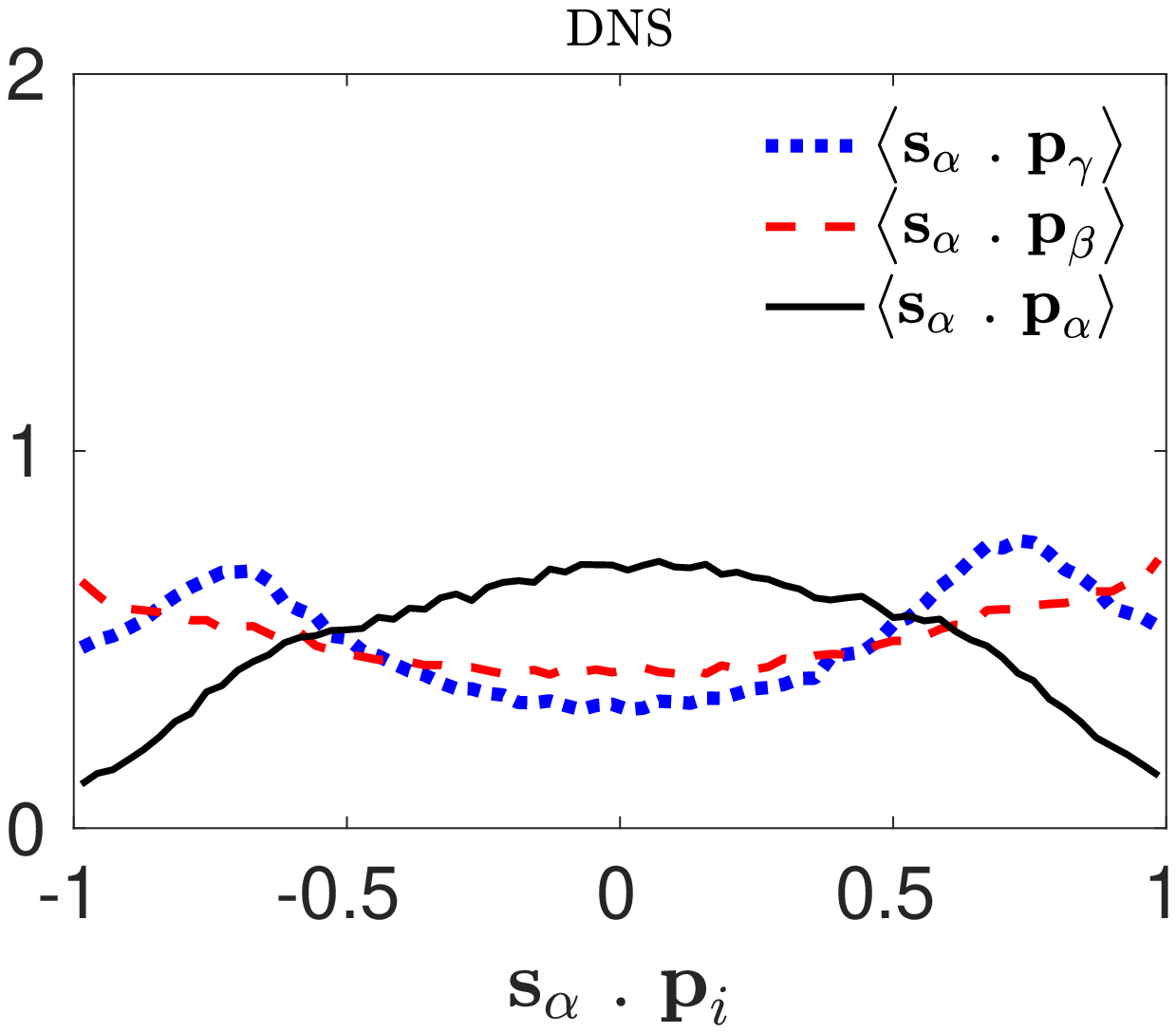}}
    \subfigure[]{
    \includegraphics[width=4.5cm]{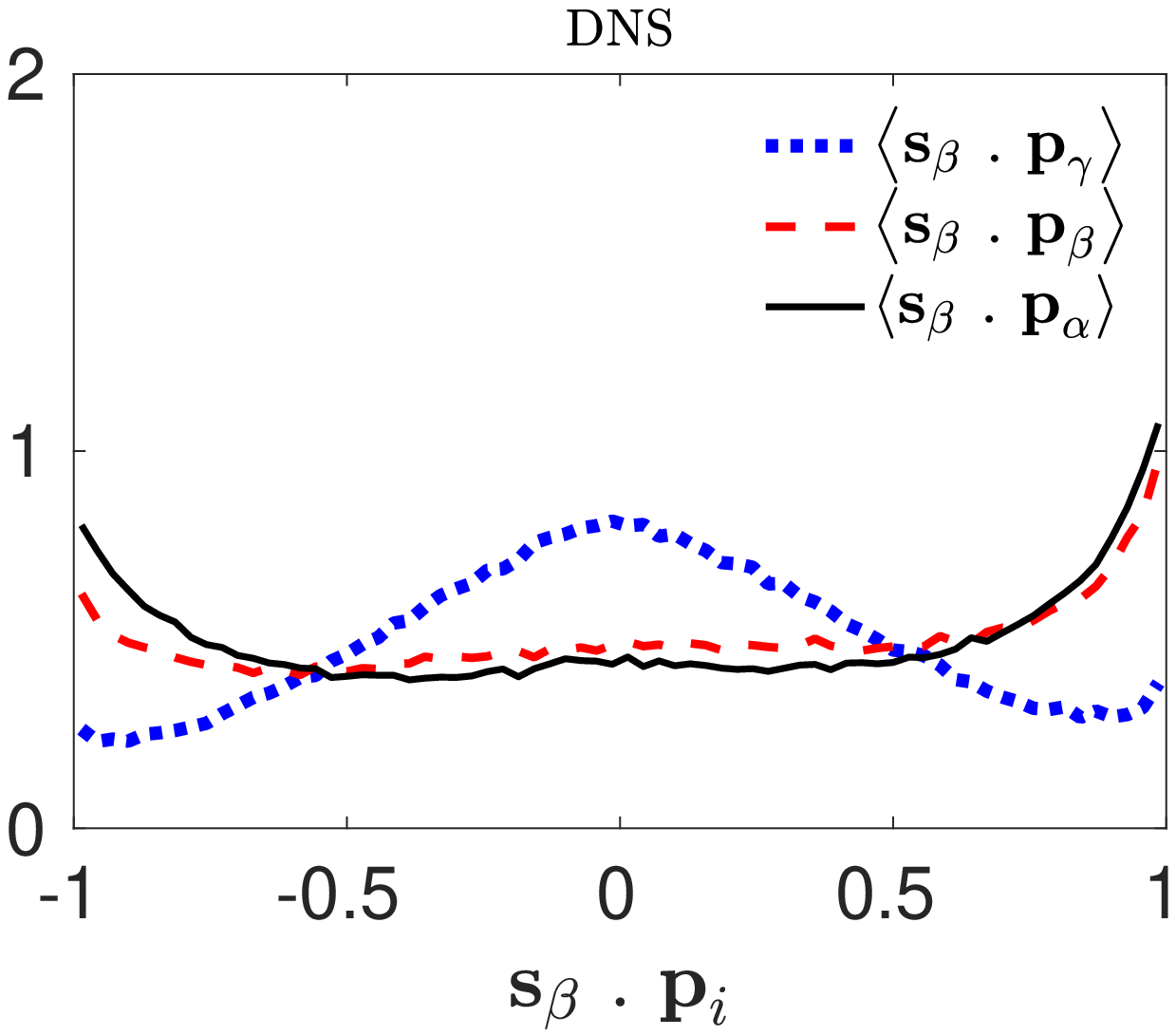}}
    \subfigure[]{
    \includegraphics[width=4.5cm]{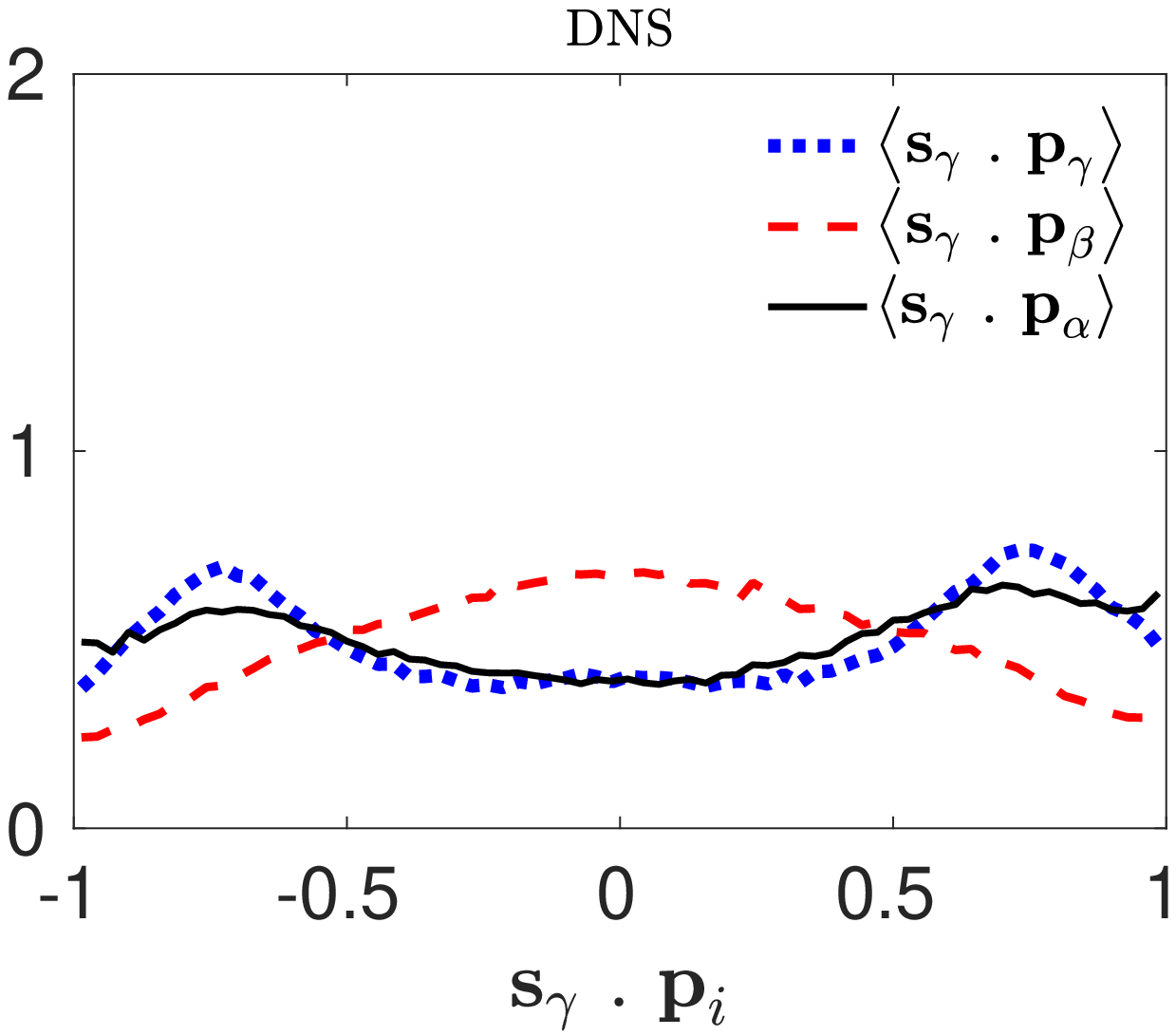}}
    \caption{Alignment of $\boldsymbol{\mathcal{P^{TBNN}}}$-eigenvectors ($\boldsymbol{p_i}$) with $\boldsymbol{S}$-eigenvectors ($\boldsymbol{s_i}$). Here, $i$ (= $\alpha$, $\beta$ or $\gamma$) denotes the three eigenvectors corresponding to the three eigenvalues $\alpha>\beta>\gamma$. (UT Texas and JHTD channel flow dataset \cite[]{JHUTD_3}, friction velocity Reynolds number of 1000.}
    \label{fig:s_dot_p_channel}
\end{figure}

The modified TBNN was trained using an isotropic turbulent flow dataset. We saw in the previous section \ref{ss:tbnn_iso} that the network can learn key features of isotropic turbulent flows, that leads to accurate predictions of pressure-Hessian especially in terms of the alignment statistics with the strain-rate eigenvectors. We, now take a step ahead and scrutinize the trained model for a different type of flow viz. the channel flow, to which the network was not exposed while training. The presence of solid walls in a channel flow leads to the generation of boundary layers near the walls. The pressure and velocity profiles in a boundary layer are very different from that observed in isotropic flow, that has no solid walls. Hence, we cannot expect our trained model to predict the pressure-Hessian for turbulent channel flow accurately. In fact, when we pass the velocity gradient information through the trained network, a very large relative Frobenius$-$norm error of 2.1838 is obtained on the predicted solution. However, the predicted output of the modified TBNN still retrieves accurate alignment statistics with the strain-rate eigenvectors, as shown in figure \ref{fig:s_dot_p_channel}. Hence, there does exist a relevant mapping between pressure-Hessian and velocity gradients that can ensure correct alignment with the strain-rate eigenvectors. The network has been able to learn this key physical mapping which is possibly independent of the type of flow and its Reynolds number (at least for isotropic and channel turbulent flows). As discussed in section \ref{s:NN}, the evolution of pressure-Hessian is expected to be governed by a large spectrum of flow quantities, their derivatives and evolution history. However, the major focus of this work has been to explore the maximum potential of local velocity gradients to describe the pressure-Hessian. We report that using only the local velocity gradient tensor we can model the pressure-Hessian such that it at least aligns with the strain-rate eigenvectors appropriately.


\subsection{Predicted coefficients by the modified network}
\begin{figure}[t]
    \centering
    \subfigure{
    \includegraphics[width=3.8cm]{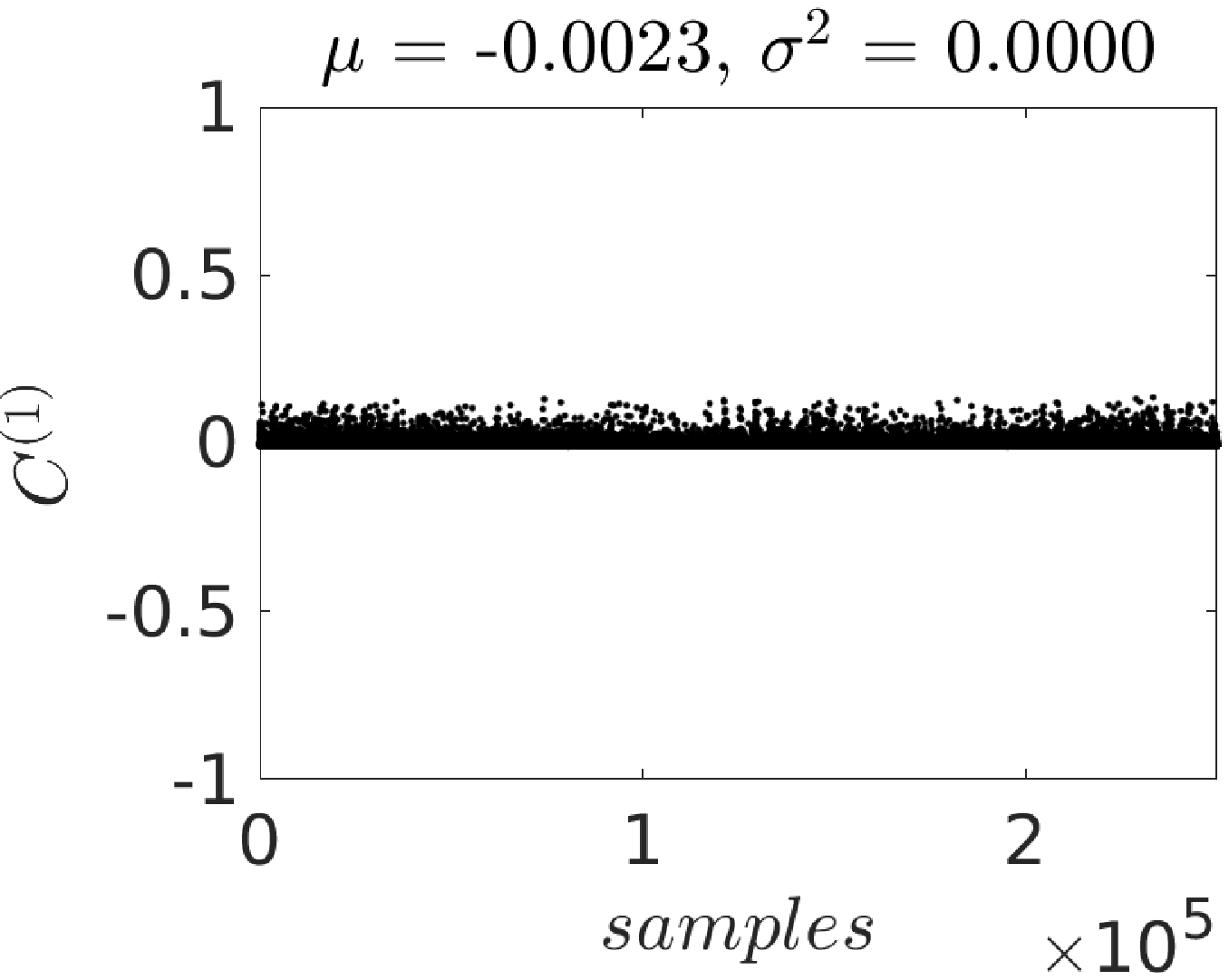}}
    \subfigure{
    \includegraphics[width=3.8cm]{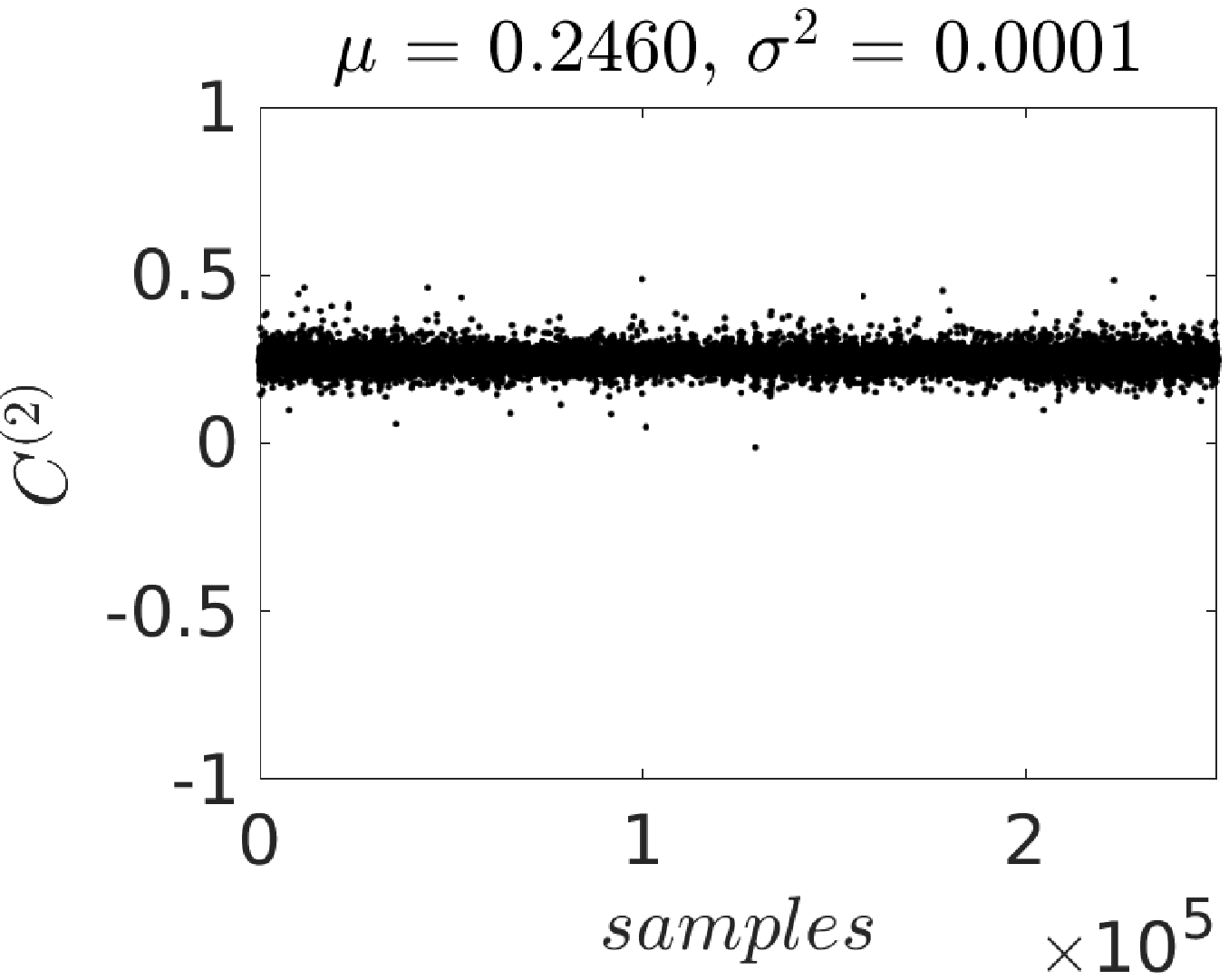}}
    \subfigure{
    \includegraphics[width=3.8cm]{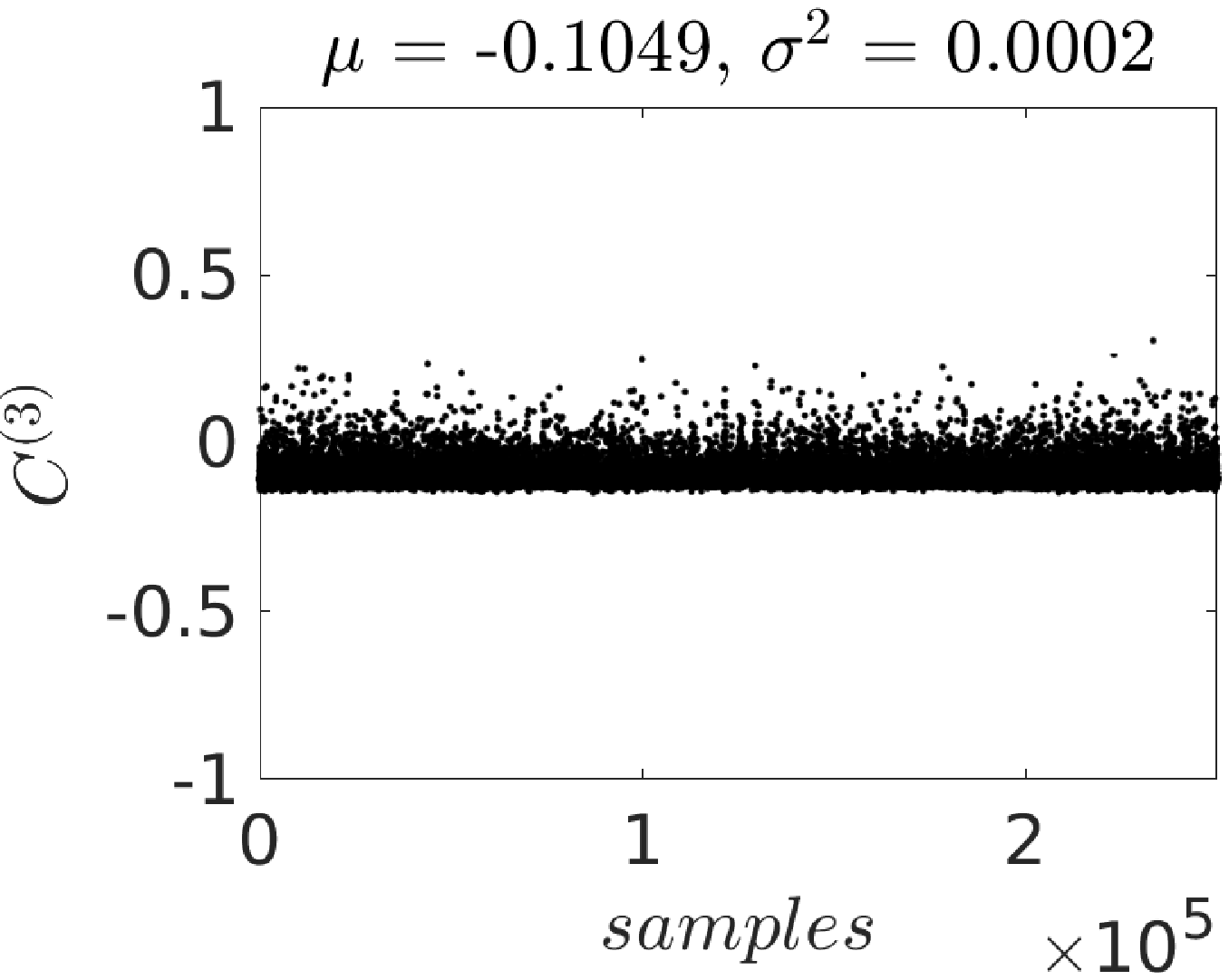}}
    \subfigure{
    \includegraphics[width=3.8cm]{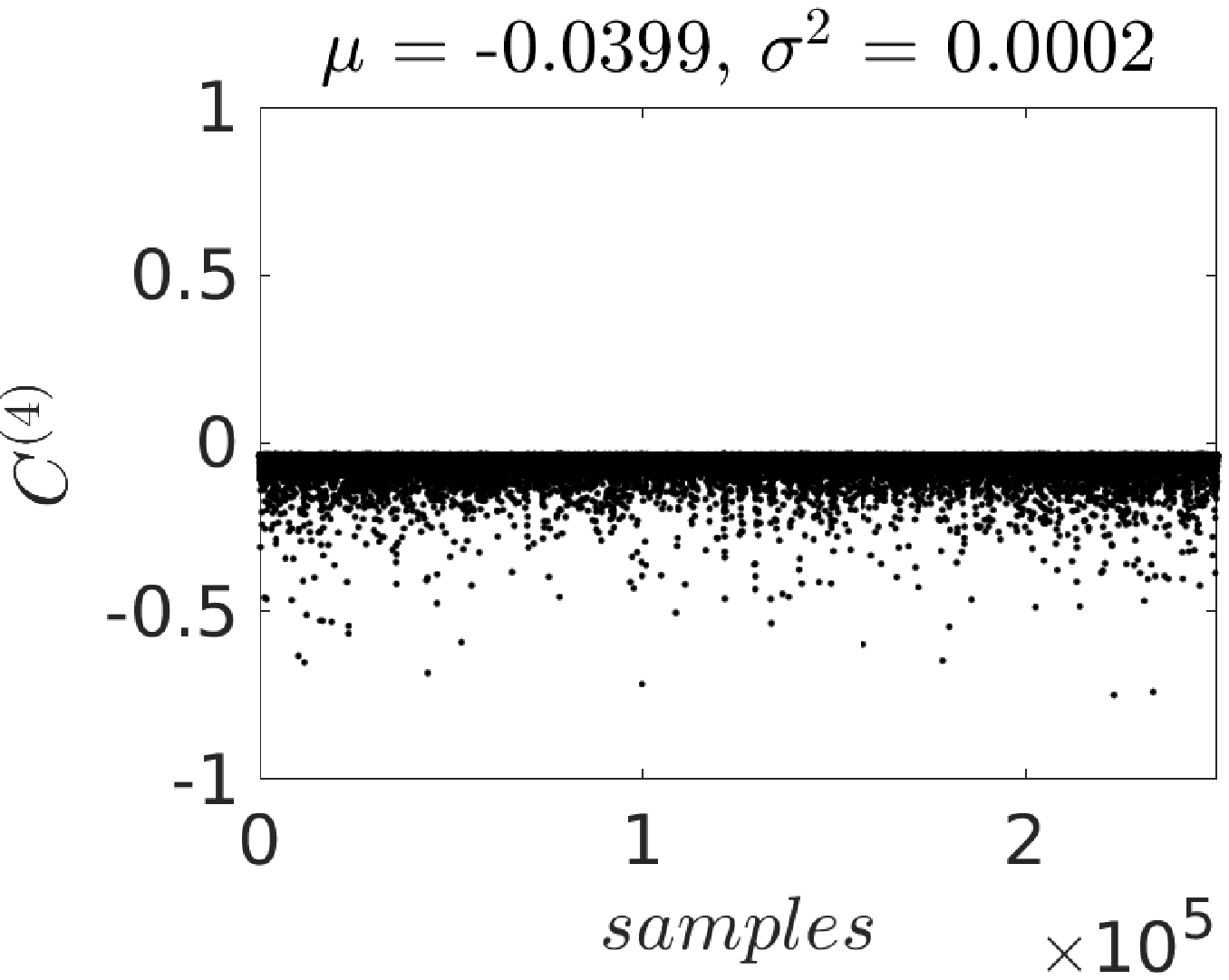}}
    \subfigure{
    \includegraphics[width=3.8cm]{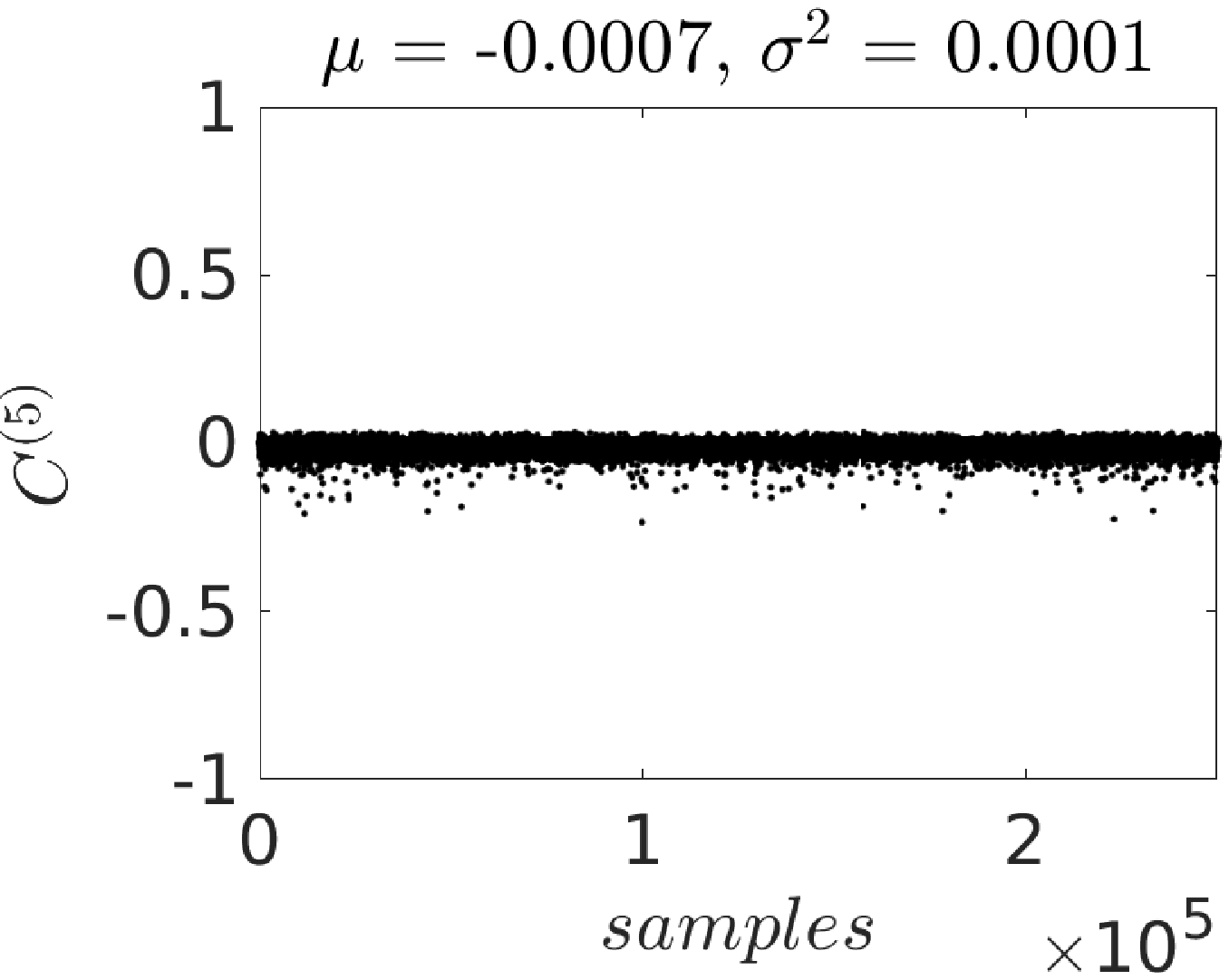}}
    \subfigure{
    \includegraphics[width=3.8cm]{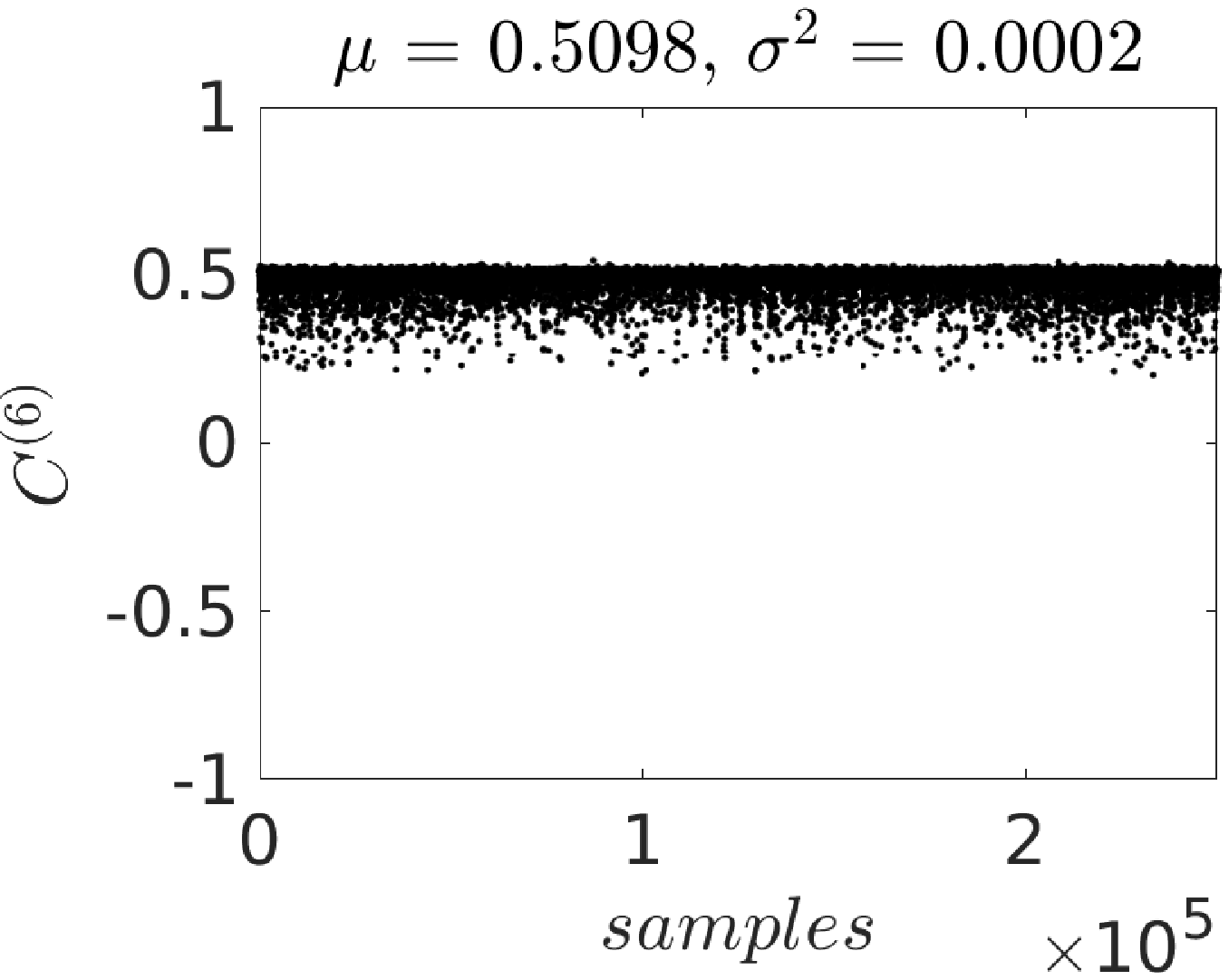}}
    \subfigure{
    \includegraphics[width=3.8cm]{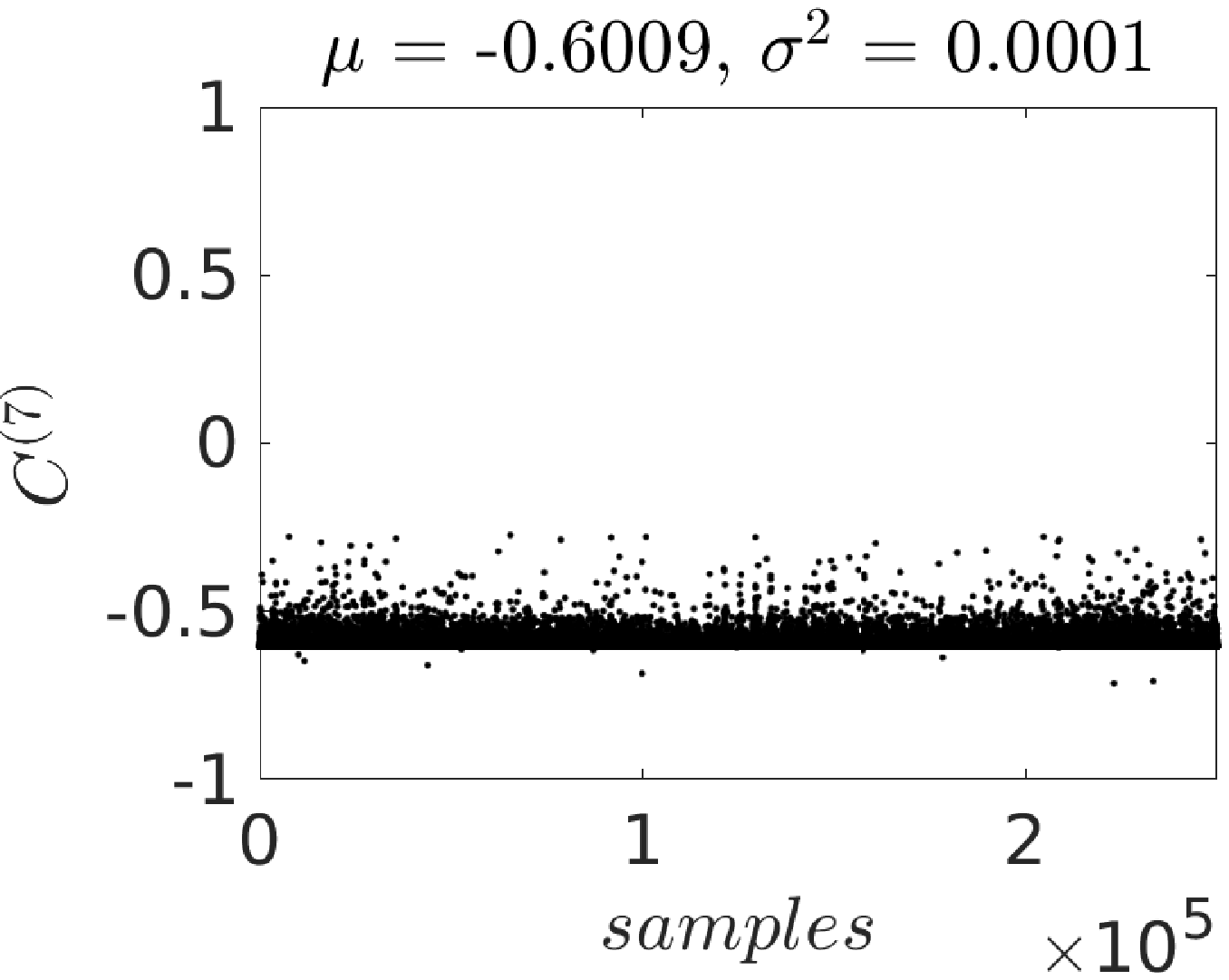}}
    \subfigure{
    \includegraphics[width=3.8cm]{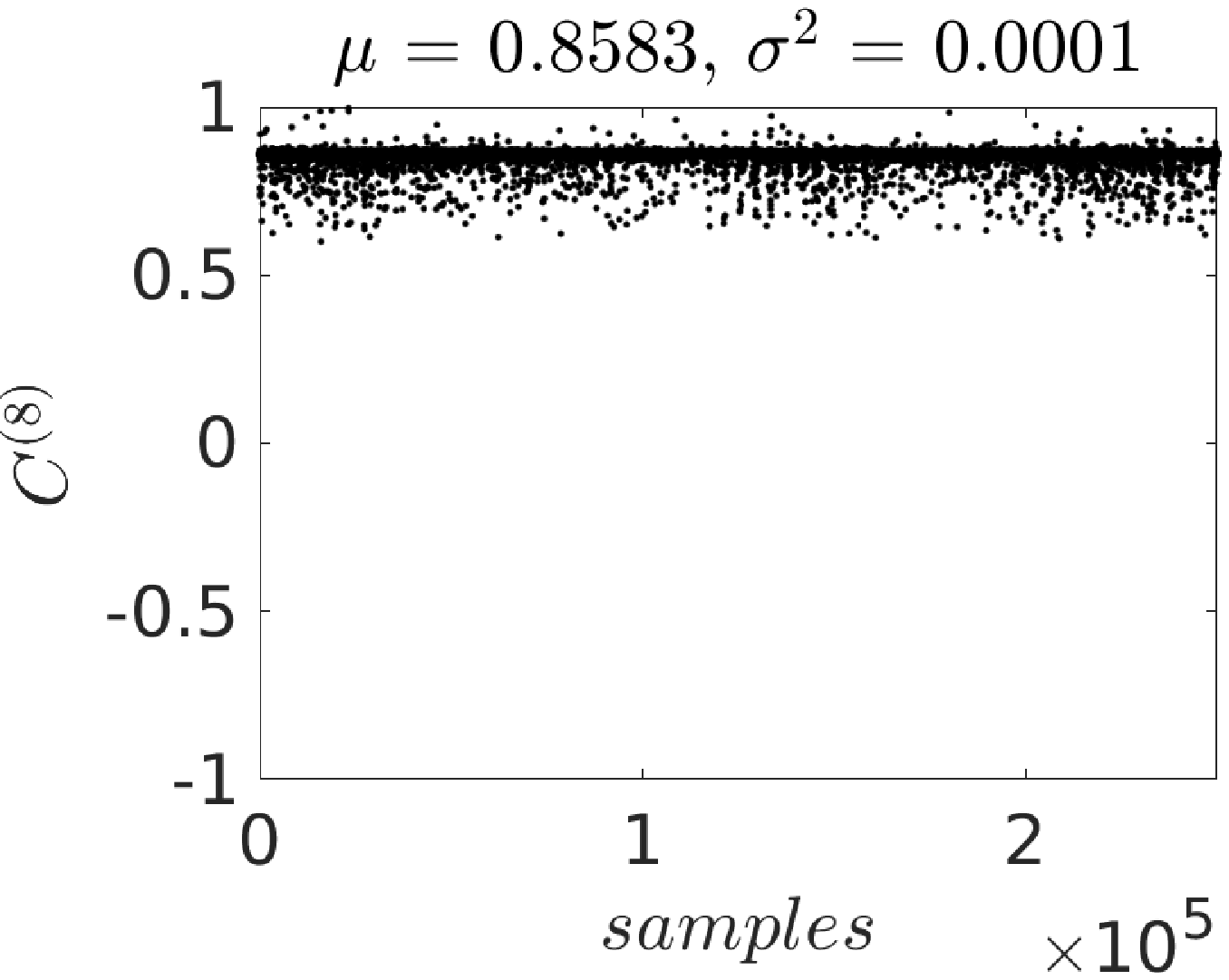}}
    \subfigure{
    \includegraphics[width=3.8cm]{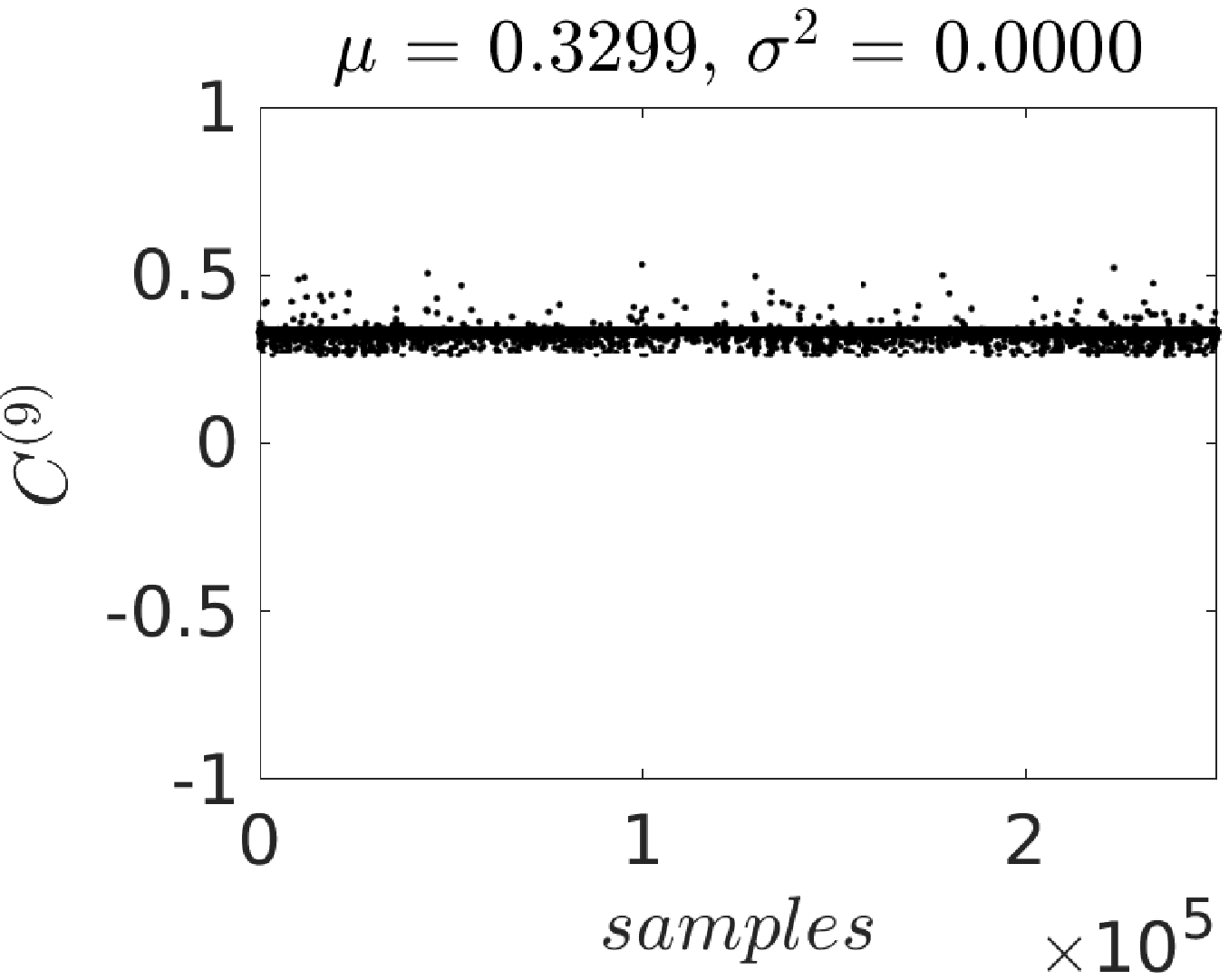}}
    \subfigure{
    \includegraphics[width=3.8cm]{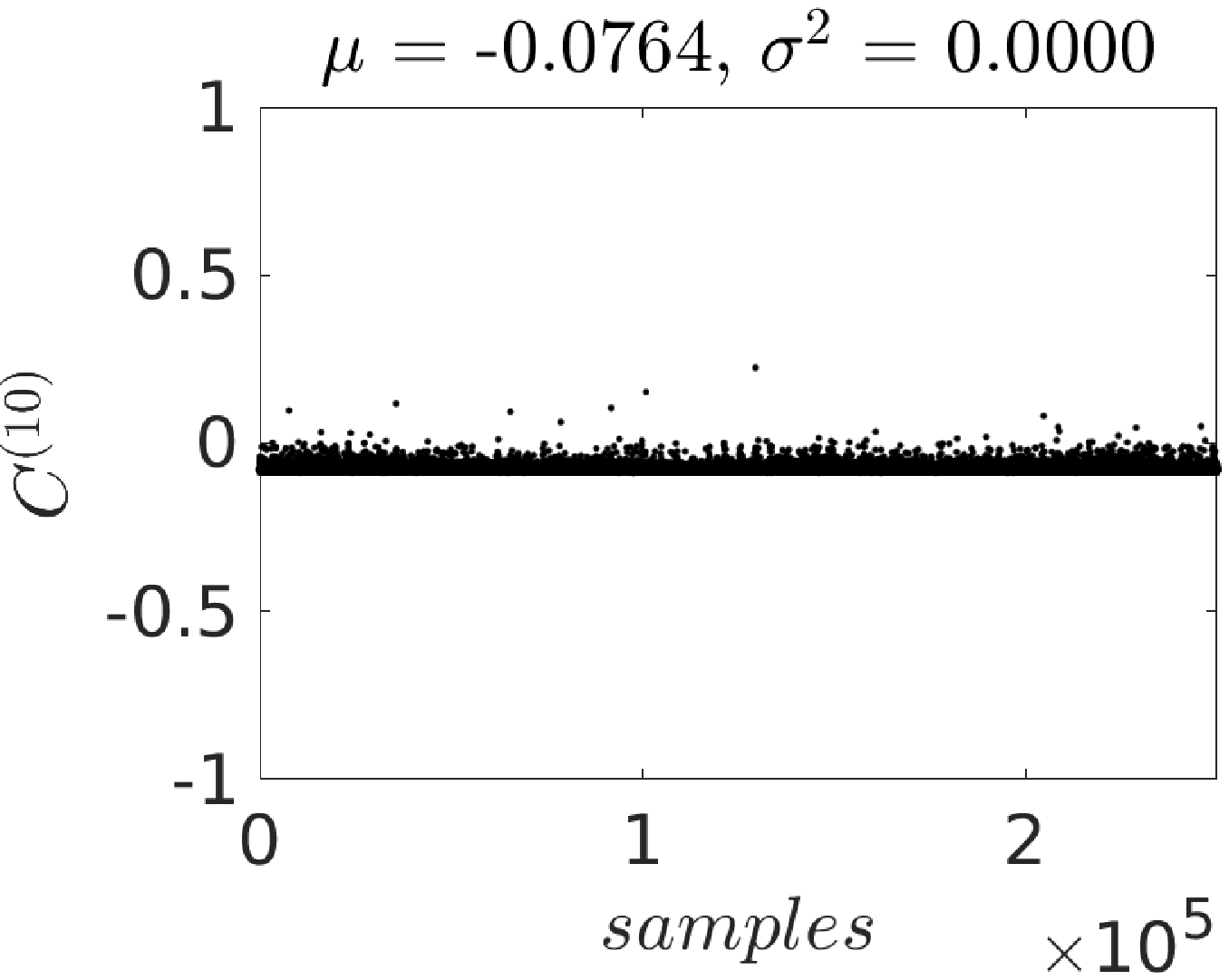}}
    \caption{Scatter plot of the ten coefficients predicted by the modified TBNN.}
    \label{fig:coeffs}
    \end{figure}

In Figure \ref{fig:coeffs}, we show the scatter plot of the coefficients predicted by the modified TBNN. We observe that each of these ten coefficients ($C^{(i)}$) have negligible variance. The overall distribution can effectively be replaced by the mean value of the distribution of each of the coefficients. Further, we find that by using the mean value of the coefficients, we retrieve the same statistics as obtained by passing the velocity gradient information through the modified TBNN. 

With this revelation, it is no longer required to use the trained network for pressure-Hessian estimation. Rather, we can just use a very simple process for pressure-Hessian prediction:
\begin{enumerate}
    \item Non-dimensionalize, the velocity gradient tensor, using the mean value of the Frobenius norm of the whole sample.
    \item Calculate the ten tensor basis and five independent invariants of strain-rate and rotation-rate tensors using equations \ref{eq:basis} and \ref{eq:invariants}. 
    \item Normalize the tensor basis using the scaling matrices used for the trained network (details in Appendix \ref{s:appendix}).
    \item Take a linear combination of the tensor basis using the mean value of the coefficients obtained from the trained network. This would yield the modelled normalized pressure-Hessian (refer Appendix \ref{s:appendix}).
    \item Scale the predicted pressure-Hessian back to its original dimensional form, using the same scaling matrices that were used while training the network.
    \item Enforce the predicted solution to have the desired trace (since the trace of $\mathcal{\boldsymbol{P}}$ is the same as the trace of $\boldsymbol{-A}^2$) 
\end{enumerate}
The complete details of the step-by-step process for calculation of the modelled pressure-Hessian tensor are presented in Appendix \ref{s:appendix}. 

\section{Conclusions}
\label{s:summary}
In this work, we first scrutinize the state of the art RFD model \cite[]{chevillard2006lagrangian} for pressure-Hessian prediction, in terms of its alignment statistics with the strain-rate eigenvectors. We report that the eigenvectors of the pressure-Hessian obtained from RFD model are either mostly parallel or perpendicular to the strain-rate eigenvectors. To decipher a better functional mapping between pressure-Hessian and velocity gradient, we employ a tensor basis neural network (TBNN) architecture \cite[]{ling2016}. The neural network is trained on high-resolution isotropic turbulence data at a Renolds number of 433. With the help of TBNN, the pressure-Hessian tensor is modelled in terms of the trace-free and symmetric tensor basis of strain-rate and rotation-rate tensors. We report that the accuracy of the predicted pressure-Hessian by TBNN is comparable to that obtained from the state of the are RFD model. However, only a marginal improvement in the alignment statistics of the TBNN output is observed. Further, we report that by scaling the tensor basis of strain-rate and rotation-rate tensors such that each element of the basis lies between $[0, 1]$, the predicted output of the neural network yields excellent alignment statistics with the strain-rate tensor for isotropic turbulent flows at different Reynolds number. Further, we test the trained model for turbulent channel flow dataset, to which the network was not exposed while training. We find that although there is significant error in element-wise comparison, the statistics of alignments obtained with the strain-rate eigenvectors are in good agreement with DNS results. With this finding, we come to the conclusion that there does exist a relevant physical mapping between pressure-Hessian and velocity gradients which enforce their eigenvectors to align appropriately with each other. This mapping is found to be independent of the type of flow and its Reynolds number (at least for isotropic turbulence and channel flow). The modified TBNN has been able to learn this key mapping by appropriately normalizing the tensor basis of strain-rate and rotation-rate tensors. Finally, we find that the distribution of the coefficients of the tensor basis obtained from the neural network has negligible variance. With this revelation, we have been able to identify ten unique coefficients of the tensor basis, the linear combination over which can be used to model the pressure-Hessian tensor directly.

\bibliography{aipsamp.bib}

\appendix
\section{}
\label{s:appendix}
We present a step-by-step process for the modelled pressure-Hessian calculation based on mean values of the coefficients derived from the modified TBNN.

\begin{enumerate}
    \item \textbf{Non-dimensionalize the strain-rate and rotation-rate tensors}
    $$\epsilon = \left< \sqrt{A_{ij}A_{ij}} \right>,$$
    where, $<>$ represents the mean over the whole sample.
    $$\textbf{S} = \frac{\textbf{A} + \textbf{A}^T}{2 \epsilon},$$
    $$\textbf{R} = \frac{\textbf{A} - \textbf{A}^T}{2 \epsilon},$$
    \item \textbf{Find the ten tensor basis and the five independent invariants}
    \begin{align}
        \boldsymbol{T}^1 &= \boldsymbol{S}, \ &\boldsymbol{T}^2 &= \boldsymbol{SR}-\boldsymbol{RS},  \nonumber\\
        \boldsymbol{T}^3 &= \boldsymbol{S}^2-\frac{1}{3}\boldsymbol{I}\{\boldsymbol{S}^2\}, \ &\boldsymbol{T}^4 &= \boldsymbol{R}^2-\frac{1}{3}\boldsymbol{I}\{\boldsymbol{R}^2\},  \nonumber\\
        \boldsymbol{T}^5 &= \boldsymbol{RS}^2-\boldsymbol{S}^2\boldsymbol{R}, \ &\boldsymbol{T}^6 &= \boldsymbol{R}^2\boldsymbol{S}+\boldsymbol{SR}^2-\frac{2}{3}\boldsymbol{I}\{\boldsymbol{SR}^2\},  \nonumber\\
        \boldsymbol{T}^7 &= \boldsymbol{RSR}^2-\boldsymbol{R}^2\boldsymbol{SR}, \ &\boldsymbol{T}^8 &= \boldsymbol{SRS}^2-\boldsymbol{S}^2\boldsymbol{RS},  \nonumber\\
        \boldsymbol{T}^9 &= \boldsymbol{R}^2\boldsymbol{S}^2+\boldsymbol{S}^2\boldsymbol{R}^2-\frac{2}{3}\boldsymbol{I}\{\boldsymbol{S}^2\boldsymbol{R}^2\},   \ &\boldsymbol{T}^{10} &= \boldsymbol{RS}^2\boldsymbol{R}^2-\boldsymbol{R}^2\boldsymbol{S}^2\boldsymbol{R}; \nonumber
    \end{align}
    \begin{align}
        \lambda^1 = \{\boldsymbol{S}^2\},  \hspace{0.65cm}  \lambda^2 = \{\boldsymbol{R}^2\},  \hspace{0.65cm}  \lambda^3 = \{\boldsymbol{S}^3\}, \hspace{0.65cm}  \lambda^4 = \{\boldsymbol{R}^2\boldsymbol{S}\}, \hspace{0.65cm}  \lambda^5 = \{\boldsymbol{R}^2\boldsymbol{S}^2\}.\nonumber
    \end{align}

    \item \textbf{Normalize the tensor basis} ($\boldsymbol{T^{(i)}}$)
    $$\boldsymbol{T}^{(i)'} = (\boldsymbol{T}^{(i)} - \boldsymbol{G}^{(i)}) \oslash (\boldsymbol{F}^{(i)} - \boldsymbol{G}^{(i)}),$$
    where, symbol $\oslash$ represents the Hadamard division between the two tensor.
    $\boldsymbol{F}^{(i)}$ and $\boldsymbol{G}^{(i)}$ represents the matrices used to scale the tensor basis $\boldsymbol{T}^{(i)}$:
            \[\boldsymbol{G}^{(1)} =
            \begin{bmatrix}
            -3.5709  & -3.0858  & -2.7680 \\
            -3.0858  & -3.0346  & -3.0692 \\
            -2.7680  & -3.0692  & -4.2937
            \end{bmatrix}, \ \  \boldsymbol{F}^{(1)} =
            \begin{bmatrix}
            2.7021  &  3.2914  &  2.3884 \\
            3.2914  &  3.3876  &  2.8077 \\
            2.3884  &  2.8077  &  2.8905
            \end{bmatrix} \]

            \[\boldsymbol{G}^{(2)} =
            \begin{bmatrix}
            -25.6104 & -31.8796 & -21.5908 \\
            -31.8796 & -30.5056 & -22.2121 \\
            -21.5908 & -22.2121 & -28.0164
            \end{bmatrix}, \ \  \boldsymbol{F}^{(2)} =
            \begin{bmatrix}
            32.0154 &  35.4956 &  20.6083 \\
            35.4956 &  29.8806 &  30.5352 \\
            20.6083 &  30.5352 &  31.7193
            \end{bmatrix} \]

            \[\boldsymbol{G}^{(3)} =
            \begin{bmatrix}
            -9.2866 &  -6.4438  & -10.5787 \\
            -6.4438 &  -6.4369  & -5.4972 \\
            -10.5787 &  -5.4972 &  -8.4430
            \end{bmatrix}, \ \  \boldsymbol{F}^{(3)} =
            \begin{bmatrix}
            8.2153  &  7.9086  &  7.8492 \\
            7.9086  &  6.0971  &  5.9908 \\
            7.8492  &  5.9908  &  9.3168
            \end{bmatrix} \]

            \[\boldsymbol{G}^{(4)} =
            \begin{bmatrix}
            -14.8512 & -15.3252 & -13.8140 \\
            -15.3252 & -11.2078 & -11.7173 \\
            -13.8140 & -11.7173 & -13.0263
            \end{bmatrix}, \ \  \boldsymbol{F}^{(4)} =
            \begin{bmatrix}
            16.6632 &   9.6823 &  12.9011 \\
            9.6823  & 27.8775  & 15.5748 \\
            12.9011 &  15.5748 &  23.0473
            \end{bmatrix} \]

            \[\boldsymbol{G}^{(5)} =
            \begin{bmatrix}
            -54.1672 & -26.9697 & -23.7045 \\
            -26.9697 & -43.5902 & -36.9865 \\
            -23.7045 & -36.9865 & -30.1412
            \end{bmatrix}, \ \  \boldsymbol{F}^{(5)} =
            \begin{bmatrix}
            27.5052 &  57.6332 &  33.7410 \\
            57.6332 &  53.5055 &  38.4958 \\
            33.7410 &  38.4958 &  55.2717
            \end{bmatrix} \]

            \[\boldsymbol{G}^{(6)} =
            \begin{bmatrix}
            -119.1511 & -78.0652 & -112.3875 \\
            -78.0652  &-186.4133 & -80.4339 \\
            -112.3875 & -80.4339 & -116.5743
            \end{bmatrix}, \ \  \boldsymbol{F}^{(6)} =
            \begin{bmatrix}
             195.0304 & 143.0951 & 108.0702 \\
             143.0951 &  92.4415 & 169.8799 \\
             108.0702 & 169.8799 & 135.2173
            \end{bmatrix} \]

            \[\boldsymbol{G}^{(7)} =
            \begin{bmatrix}
            -737.6056  & -1038.4397 & -640.3922 \\
            -1038.4397 & -963.7453  & -551.0073 \\
            -640.3922  & -551.0073  & -537.7603
            \end{bmatrix}, \ \  \boldsymbol{F}^{(7)} =
            \begin{bmatrix}
            780.8498 & 970.6533 & 540.2198 \\
            970.6533 & 808.6133 & 523.5790 \\
            540.2198 & 523.5790 & 924.0598
            \end{bmatrix} \]

            \[\boldsymbol{G}^{(8)} =
            \begin{bmatrix}
            -496.1792 & -376.4132 & -245.1099 \\
            -376.4132 & -341.9391 & -345.2818 \\
            -245.1099 & -345.2818 & -405.4714
            \end{bmatrix}, \ \  \boldsymbol{F}^{(8)} =
            \begin{bmatrix}
            390.7658 & 778.0012 & 298.8682 \\
            778.0012 & 658.3773 & 509.5283 \\
            298.8682 & 509.5283 & 444.1750
            \end{bmatrix} \]

            \[\boldsymbol{G}^{(9)} =
            \begin{bmatrix}
            -526.6148 & -345.7790 & -276.1035 \\
            -345.7790 & -188.0288 & -293.5551 \\
            -276.1035 & -293.5551 & -339.0242
            \end{bmatrix}, \ \  \boldsymbol{F}^{(9)} =
            \begin{bmatrix}
            381.9239 & 194.5271 & 578.1185 \\
            194.5271 & 197.7692 & 170.3597 \\
            578.1185 & 170.3597 & 598.0897
            \end{bmatrix} \]

            \[\boldsymbol{G}^{(10)} =
            \begin{bmatrix}
            -595.3592  & -1679.5692  & -922.0738 \\
            -1679.5692 &  -1466.6689 & -674.1032 \\
            -922.0738  &  -674.1032  & -1079.9475
            \end{bmatrix}, \ \  \boldsymbol{F}^{(10)} =
            \begin{bmatrix}
            1335.8832 &  1173.0048 &  438.1600 \\
            1173.0047 &  1126.0351 &  671.8583 \\
            438.1600  &  671.8582  &  662.4745
            \end{bmatrix} \]

    \item \textbf{Take a linear combination of tensor basis using mean coefficient values}
    \begin{equation}
        \boldsymbol{\mathcal{P}^{TBNN'}} = \sum_{i=1}^{10} C^i \boldsymbol{T}^{i'}, \nonumber
    \end{equation}
    where, $C^i$ are the mean coefficient values predicted by the modified TBNN:
    \begin{align}
    C^{(1)}  &= -0.0023, \ \ C^{(2)}  =  +0.2460, \ \ C^{(3)}  &= -0.1049, \ \ C^{(4)}  &= -0.0400, \ \ C^{(5)}  &= -0.0007  \nonumber \\
    C^{(6)}  &=  +0.5098, \ \ C^{(7)}  = -0.6009, \ \ C^{(8)}  &=  +0.8583, \ \ C^{(9)}  &=  +0.3299, \ \ C^{(10)} &= -0.0764 \nonumber
    \end{align}

    \item \textbf{Scale the predicted pressure-Hessian back to its dimensional form}
    \begin{equation}
        \boldsymbol{\mathcal{P}^{TBNN'}} = \left[\boldsymbol{\mathcal{P}^{TBNN'}} \circ (\boldsymbol{F_p} - \boldsymbol{G_p}) + \boldsymbol{G_p}\right] \epsilon^2, \nonumber
    \end{equation}
    where, the symbol $\circ$ represents the Hadamard product of two matrices and  $\boldsymbol{G_p}$ and $\boldsymbol{F_p}$ are the scaling matrices:
    
            \[\boldsymbol{G}_p =
            \begin{bmatrix}
            -26.9693 & -13.2321 & -12.5971 \\
            -13.2321 & -26.0595 & -17.4419 \\
            -12.5971 & -17.4419 & -24.2304
            \end{bmatrix}, \ \  \boldsymbol{F}_p =
            \begin{bmatrix}
            19.1816 &  23.7427 &  17.5106 \\
            23.7427 &  27.9531 &  25.3751 \\
            17.5106 &  25.3751 &  20.1042
            \end{bmatrix} \]

    \item \textbf{Trace correction step}
    \begin{equation}
        \boldsymbol{\mathcal{P}^{TBNN}} = \boldsymbol{\mathcal{P}^{TBNN'}} - \{\boldsymbol{\mathcal{P}^{TBNN'}}\} \frac{\textbf{I}}{3} + \{\boldsymbol{-A}^2\} \frac{\textbf{I}}{3}, \nonumber
    \end{equation}
    where, $\textbf{I}$ is the identity matrix and $\{\}$ represents the trace of the matrix.
    
\end{enumerate}

\end{document}